\documentclass[lettersize,journal]{IEEEtran}
\usepackage{amsmath,amsfonts}
\usepackage{algorithmic}
\usepackage{algorithm}
\usepackage{array}
\usepackage[caption=false,font=normalsize,labelfont=sf,textfont=sf]{subfig}
\usepackage{textcomp}
\usepackage{stfloats}
\usepackage{url}
\usepackage{verbatim}
\usepackage{graphicx}
\usepackage{cite}

\usepackage{diagbox} 
\usepackage{multirow} 
\usepackage{booktabs} 
\usepackage{hyperref}
\hypersetup{
    colorlinks = true,
    linkcolor = blue,
    urlcolor = blue,
    citecolor = blue,
}

\begin{document}

\title{Single-Domain Generalized Object Detection by Balancing Domain Diversity and Invariance}

\author{Zhenwei He~\IEEEmembership{IEEE Member}, Hongsu Ni
\thanks{This paper was produced by the IEEE Publication Technology Group. They are in Piscataway, NJ.}
\thanks{Manuscript received April 19, 2021; revised August 16, 2021.}}

\markboth{Journal of \LaTeX\ Class Files,~Vol.~14, No.~8, August~2021}%
{Shell \MakeLowercase{\textit{et al.}}: A Sample Article Using IEEEtran.cls for IEEE Journals}

\IEEEpubid{0000--0000/00\$00.00~\copyright~2025 IEEE}

\maketitle

\begin{abstract}

Single-domain generalization for object detection (S-DGOD) seeks to transfer learned representations from a single source domain to unseen target domains. While recent approaches have primarily focused on achieving feature invariance, they ignore that domain diversity also presents significant challenges for the task. First, such invariance-driven strategies often lead to the loss of domain-specific information, resulting in incomplete feature representations. Second, cross-domain feature alignment forces the model to overlook domain-specific discrepancies, thereby increasing the complexity of the training process. To address these limitations, this paper proposes the Diversity Invariant Detection Model (DIDM), which achieves a harmonious integration of domain-specific diversity and domain invariance. Our key idea is to learn the invariant representations by keeping the inherent domain-specific features. Specifically, we introduce the Diversity Learning Module (DLM). This module limits the invariant semantics while explicitly enhancing domain-specific feature representation through a proposed feature diversity loss. Furthermore, to ensure cross-domain invariance without sacrificing diversity, we incorporate the Weighted Aligning Module (WAM) to enable feature alignment while maintaining the discriminative domain-specific information. Extensive experiments on multiple diverse datasets demonstrate the effectiveness of the proposed model, achieving superior performance compared to existing methods.
\end{abstract}


\section{Introduction}
\label{sec1}
\IEEEPARstart{I}{n} recent years, deep learning has led to significant improvements in the performance of detection models \cite{ren2016faster,tian2019fcos,ge2021yolox,wu2021generalized}. However, traditional detectors often rely on the assumption of Independent and Identically Distributed (i.i.d.) data. This assumption becomes a limitation when the model encounters a domain shift \cite{chen2020harmonizing,torralba2011unbiased}, leading to a significant drop in performance and restricting the applicability of these models in real-world scenarios.

To address the domain shift challenge, domain adaptation (DA) methods were developed with the objective of improving model performance in the target domain by minimizing the domain discrepancy between source and target domains. However, the unavailability of the true label of the target domain has led researchers to indirectly manage target risk through source domain optimization while explicitly accounting for domain distributional differences \cite{wang2022generalizing}. According to the theoretical framework established by Ben-David et al. \cite{ben2010theory}, DA typically employs feature alignment strategies to reduce the statistical divergence between domains. This approach, grounded in domain-invariant representation learning \cite{ganin2016domain,courty2016optimal,nguyen2021domain,hou2025gradient}, has become a central research direction in domain adaptation and domain generalization, simultaneously driving methodological innovations in these fields.

\begin{figure*}
    \centering
    \includegraphics[width=0.95\linewidth]{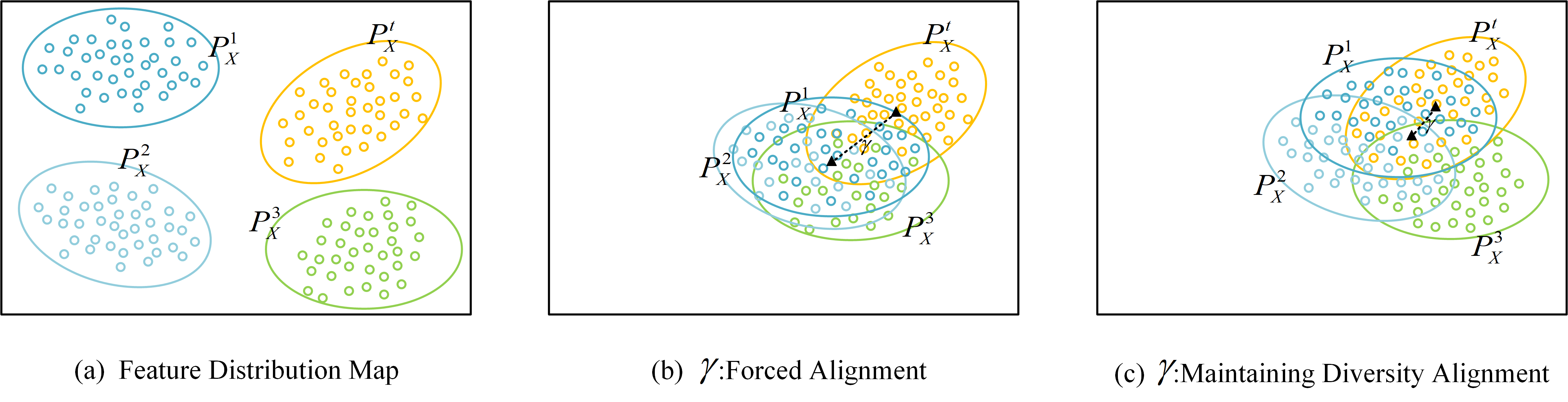}
    \caption{Illustrate three scenarios of feature distribution: (a) unaligned features, (b) fully aligned features, (c) aligned features with domain-specific information preserved. The essence of feature alignment is to minimize the difference in distribution $\gamma$ between the source domain combination and the target domain. However, as illustrated in Fig. (b), forced alignment may result in an overly homogeneous feature space, neglecting the unique structural information inherent to each domain and failing to effectively cover the feature distribution of the target domain. Therefore, it is essential to consider the diversity of domain-specific information. As demonstrated in Fig. (c), this diversity-protecting alignment strategy can create a more inclusive and representative feature space, significantly enhancing the overlap between the distribution of combined features and the target domain.}
    \label{fig:enter-label1}
\end{figure*}

Building upon domain adaptation (DA) methods, existing domain generalization (DG) models also aim to reduce domain bias by learning domain-invariant features. By minimizing the risk across multiple source domains, these models extract features that are universally applicable to all source domains, thereby enhancing their generalization capability to unseen target domains. However, these approaches face three key limitations: information loss, conflicting constraints, and theoretical imbalances. Firstly, an overemphasis on invariance may hinder the model's ability to capture domain-specific information, thereby compromising the completeness of feature representation. Secondly, this approach lacks rationality, as it requires the model to extract similar or even identical features from source domains that exhibit significant visual or semantic differences. Given the inherent diversity across domains, such constraints are counterintuitive and may hinder effective feature extraction, potentially complicating the training process. Finally, as illustrated in Fig. \ref{fig:enter-label1}, the feature distributions of three scenarios are presented: (a) unaligned features, (b) fully aligned features without considering domain-specific characteristics, and (c) aligned features with domain-specific information preserved. Unlike traditional DA methods, where target domain samples are available for alignment, DG faces the challenge of unseen target domains. Simply aligning features from multiple source domains may overly constrain the feature distribution, leading to limited overlap with the probable target domain distribution. As shown in Fig. \ref{fig:enter-label1}(c), the DG task must also account for domain-specific information to enable the model to achieve greater alignment with the target domain. To address these challenges, we propose that an effective DG model should synergistically optimize the learning of domain-invariant features while preserving and leveraging domain-specific diversity. Learning domain-invariant features ensures robustness across diverse domains, whereas fully exploiting domain-specific diversity significantly enhances feature representation and optimizes the training process. The ultimate goal is to construct a more inclusive and representative feature space that maximizes overlap with the target domain distribution.


Based on the above discussion, specifically, in this paper, we introduce the Diversity Invariance Domain Generalization Detection Model (DIDM), designed to not only learn the domain invariance but also keep the diversity domain-specific. To manage these two feature types, we incorporate the Dive-\\ \\
rsity Learning Module (DLM) to capture domain-specific features and the Weighted Aligning Module (WAM) to focus on domain-invariant features.

The Diversity Learning Module (DLM) is specifically designed to preserve the diversity of domain-specific features during the training phase. First, to enhance the distinctiveness of domain-specific features, the module explicitly separates domain-invariant information from domain-specific characteristics. Specifically, since domain-specific features should not contain common semantics across domains, we propose to suppress shared semantic representations by estimating domain-specific category semantics. This mechanism ensures that the learned features retain unique domain characteristics. Second, to further enhance the diversity of domain-specific features, we propose a feature diversity loss function. This loss explicitly encourages the model to capture and enrich the representational diversity of domain-specific features during training. By separating and enriching the domain-specific features, our model can learn more discriminative features, leading to an expanded and more robust feature distribution for the target domain.

In addition to preserving the diversity of domain-specific features, achieving domain-invariant representations is also critical for domain generalization (DG) tasks. However, overemphasizing domain-invariant features may also lead to the loss of discriminative domain-specific characteristics, thereby harming the overall feature diversity. To address this challenge, we introduce the Weighted Aligning Module (WAM) to achieve feature alignment while maintaining the inherent diversity of domain-specific features. Considering that excessive alignment often causes the collapse of feature diversity, we propose that features from different domains should retain a certain level of discrepancy. Specifically, we argue that if the features can achieve similar detection performance, further alignment may not be necessary. To this end, the WAM incorporates an adaptive loss weighting mechanism that dynamically balances the trade-off between alignment and diversity. This mechanism prevents the model from overemphasizing feature alignment and instead focuses on the most relevant features for alignment.
By co-optimizing the Diversity Learning Module (DLM) and the WAM, our model achieves effective domain generalization on detection tasks. In conclusion, the contributions of the proposed Domain-Invariant Diversity Module (DIDM) are summarized as follows:

\begin{itemize} 
\item We propose the Diversity Learning Module (DLM) to preserve the inherent feature diversity within domain-specific representations. Additionally, this paper analyze the theoretical significance of feature diversity in domain generalization (DG) tasks through the transfer theory.

\item The Weighted Aligning Module (WAM) is designed to achieve cross-domain feature alignment while maintaining the intrinsic diversity of domain-specific representations. 

\item Extensive experimental evaluations on diverse benchmark datasets demonstrate the robustness and generalization capability of our proposed method, validating its effectiveness across multiple scenarios. 
\end{itemize}


\section{Related Work}
\label{sec:Work}


\textbf{Domain Generalization (DG).} Researchers have been working extensively in single domain generalization and have developed numerous methods to enhance it \cite{dou2019domain, liu2024unbiased, li2024prompt, wu2024g, zhou2022domain}. These methods can generally be categorized into three main groups: data augmentation, presentation learning of domain invariance, and various learning strategies. First, the data argumentation approach \cite{lee2022wildnet,li2023deep,somavarapu2020frustratingly} increases the diversity of source domain data by applying various data manipulation techniques. For instance, Somavarapu et al. \cite{somavarapu2020frustratingly} introduce a straightforward image stylization transform to generate diverse samples, exploring variability across sources. Similarly, Wang et al. \cite{wang2021learning} employ adversarial training to create varied input images, further enhancing generalization. Second, some kinds of DG methods focus on the presentation learning of domain invariance. Shao et al. \cite{shao2019multi} use multi-adversarial discriminative training to extract both shared and distinctive feature representations across multiple source domains. Last, Various learning strategies \cite{chen2023meta,du2020learning,peng2022semantic,seo2020learning,wang2023improving,zhao2021learning} are employed to improve model generalization. For example, Zhao et al. \cite{zhao2021learning} apply a meta-learning approach that simulates the model’s adaptation to new, unseen domains during training, thereby boosting its adaptability in unfamiliar environments. These approaches illustrate significant advancements in domain generalization, highlighting different techniques designed to improve models' performance on previously unseen domains.

\textbf{Object Detection.} 
To enhance the model's performance in object detection, researchers have conducted numerous experiments and proposed various methods in object detection. One prominent approach is Domain Adaptive Object Detection (DAOD) \cite{chen2021i3net,deng2021unbiased,krishna2023mila,li2022cross,oza2023unsupervised,zhao2020review}, which enhances the model's robustness in new domains by simultaneously training on both source and target domains. For instance, a common strategy involves minimizing the distance between global and local features, as outlined in \cite{cao2023contrastive,chen2020harmonizing,deng2021unbiased,saito2019strong}. However, these methods require access to the target domain during model training, which imposes limitations on practical applications. Therefore, researchers have proposed Domain Generalized Object Detection (DGOD) \cite{chang2024unified,qin2024towards}. For instance, Chang et al. \cite{chang2024unified} enhance the model's generalization ability by decoupling depth estimation from dynamic perspective enhancement. However, the effectiveness of DGOD largely depends on the number of accessible source domains, and collecting multiple source domains can be costly. Consequently, researchers have introduced the more challenging Single Domain Generalized Object Detection (S-DGOD) \cite{vidit2023clip,danish2024improving,lee2024object,pan2018two,choi2021robustnet}, which is generally categorized into feature normalization and invariant-based methods. Firstly, feature normalization methods such as IBN-Net, proposed by Pan et al. \cite{pan2018two}, enable the network to adjust its normalization strategy according to different tasks and datasets by combining Instance Normalization (IN) and Batch Normalization (BN). IterNorm, introduced by Huang et al. \cite{huang2019iterative}, avoids feature decomposition through a Newton iteration method, thereby improving the efficiency of the normalization process. Secondly, invariant-based approaches, such as UFR proposed by Liu et al. \cite{liu2024unbiased}, enhance the model's generalization performance by eliminating prototype bias and attentional bias. Wu et al. \cite{wu2022single} propose a cyclic disentanglement self-distillation approach specifically for single-domain generalization in object detection, which enhances feature disentanglement. These approaches demonstrate significant progress in the field of target detection and contribute to the rapid advancement of the discipline.

\section{Methodology}
\label{sec:method}
In this section, we introduce the proposed Diversity Invariance Detection Model (DIDM), which comprises two key components: the Diversity Learning Module (DLM) and the Weighted Aligning Module (WAM). These components are synergistically designed to balance domain-invariant feature learning and feature diversity. Specifically, the DLM is engineered to preserve the intrinsic diversity of domain-specific features. Meanwhile, the WAM facilitates cross-domain feature alignment to promote domain invariance.

\subsection{Preliminaries}
\textbf{Problem Description.} 
For single-domain generalization detection tasks, the model is trained using one source domain $D_s=\{{(x^i_s,y^i_s,b^i_s)}\}^{N_s}_{i=1}$, where $N_s$ stand for the number of samples,  $x^i_s$ presents the input image, $b^i_s$ and $y^i_s$ are the ground truth bounding boxes and corresponding labels, respectively. The model is trained for the detection of the unseen target domain $D_t=\{{(x^i_t)}\}^{N_t}_{i=1}$. Note that the source and target domains share the same label space.

\textbf{Theoretical Analysis.}
\label{Proof}
To address domain bias and optimize risk $\xi^t(h)$ in the target domain, domain adaptation typically aligns the probability distributions of the source $P_X^s$ and target $P_X^t$ domains while also constraining the risk $\xi^s(h)$ associated with the source domain. This mechanism is theoretically grounded in the Ben-David generalization bound \cite{ben2010theory,wang2022generalizing}.
\begin{equation}
\xi^t(h)\leq \xi^s(h)+d_{\mathcal{H}\Delta \mathcal{H}}(P_X^s,P_X^t)+\lambda_ \mathcal{H}
\end{equation}
Where $\lambda_ \mathcal{H} := inf_{h\in \mathcal{H}}[\xi^s(h)+\xi^t(h)]$ indicates the ideal joint risk and $h$ represents an arbitrary classifier within the hypothesis space. $d_{\mathcal{H}\Delta \mathcal{H}}(P_X^s,P_X^t):=sup_{h,h' \in \mathcal{H}}|\xi^s(h,h')-\xi^t(h,h')|$ is utilized to measure inter-domain distributional differences. With domain adaptation theory, many domain adaptation methods aim to minimize the domain divergence $d_{\mathcal{H}\Delta \mathcal{H}}(P_X^s,P_X^t)$ through feature alignment, achieving great performance. These approaches have been increasingly adopted in domain generalization tasks in recent years, demonstrating their effectiveness in bridging distributional gaps across heterogeneous domains.

However, the theoretical foundation of domain generalization models differs from that of domain adaptation frameworks. While DA primarily focuses on reducing domain discrepancy between source and target domains, DG theory emphasizes the broader challenge of not only domain discrepancy but also domain diversity. Specifically, the theory domain generalization is presented as follows \cite{albuquerque2019adversarial,wang2022generalizing}:

\begin{equation}
\xi^t(h)\leq \sum \limits_{i=1}^M\pi_i^*\xi^i(h)+\frac{\gamma+\rho}{2}+\lambda_{H,(P_X^t,P_X^*)}
\end{equation}
\begin{equation}
\gamma:=min_{\pi \in \Delta M}d_ \mathcal{H}(P_x^t,\sum\limits_{i=1}^M \pi_i P_X^i)
\end{equation}
\begin{equation}
\rho:=sup_{P_X',P_X'' \in \delta}d_ \mathcal{H}(P_X',P_X'')
\end{equation}
\begin{equation}
\delta:=\{\sum \limits_{i=1}^M \pi_i P_X^i | \pi \in \Delta_M\}
\end{equation}
Here, $M$ denotes the number of source domains, $P_X$ represents the feature distribution of a source domain, $\rho$ is the diameter of $\delta$, and $\gamma$ indicates the distance between $P_x^t$ and the joint feature distribution $\delta$, $P_X^*$ is the best approximating distribution. 

The theory reveals that two key factors influence the target error bound in domain generation task: (1) the domain discrepancy among source domains ($\rho$), and (2) the distance between the joint feature distribution $\delta$ and the target domain ($\gamma$). Conventional methods primarily focus on minimizing the domain discrepancy $\rho$ by aligning source domains. However, this approach may inadvertently reduce the coverage of the joint feature distribution $\delta$, thereby increasing the distance between the joint distribution and the target domain. This trade-off can degrade generalization performance on unseen target domains and make the model to achieve suboptimal performance. To address this, we propose the DIDM framework, which simultaneously optimizes two objectives: (1) learning domain-invariant features to reduce $\rho$, and (2) enhancing the diversity of domain-specific information to expand the coverage of $\delta$. By balancing these goals, our method reduces both $\gamma$ and $\rho$, ultimately tightening the target error bound and improving model robustness to unseen domains.


\begin{figure*}
    \centering
    \includegraphics[width=0.95\linewidth]{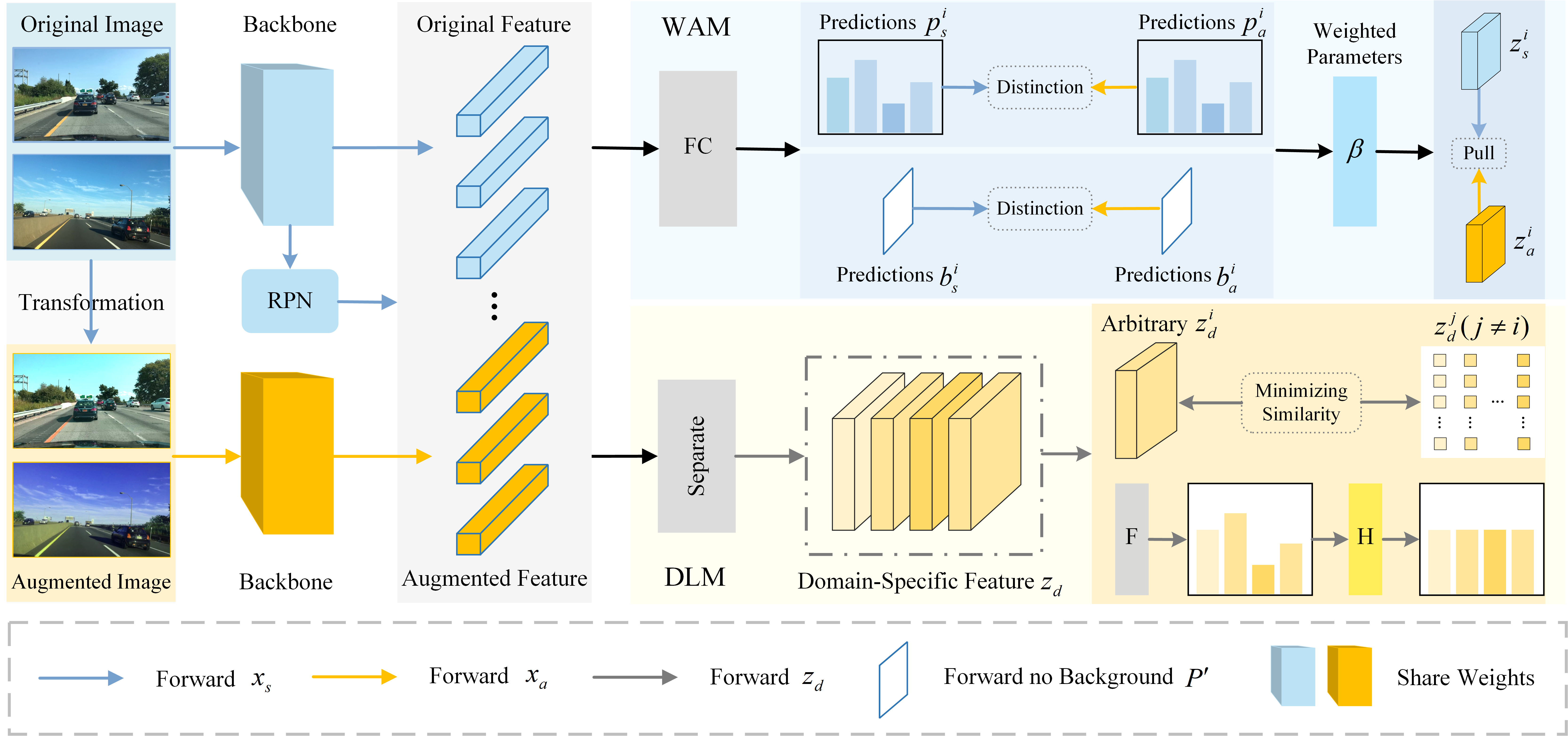}
    \caption{The general architecture of DIDM. The input image is processed through the backbone to extract both original and enhanced features, as well as to obtain the objective domain-specific feature. Subsequently, this feature is processed through the Diversity Learning Module (DLM) and the Weighted Aligning Module (WAM). DLM is implemented for domain-specific features with maximum entropy loss and feature diversity loss, where the diversity is enhanced and semantic information is suppressed. Weighted Aligning Module (WAM) imposes a constraint on feature alignment, which prevents the model's overemphasis on the alignment and compromises diversity learning. Together, the DLM and WAM work synergistically to ensure that the model focuses on robust features.}
    \label{fig:enter-label2}
\end{figure*}
\subsection{Overview}

The net structure of the proposed model is shown in Fig. \ref{fig:enter-label2}, where Faster R-CNN \cite{ren2016faster} is used as the base detector. Given a source image $x_s$, we can generate an argument image $x_a=A(x_s)$. During the training phase, both $x_s$ and $x_a$ are fed into the model to produce their corresponding feature maps $F_s$ and $F_a$, respectively. Only the $F_s$ is fed into RPN to get a series of proposals $P \in R^{K \times 4}$. To suppress background influence, background boxes are filtered from the proposals $P$, retaining only the foreground boxes $P'$. Then, the $F_s$, $F_a$ and $P'$ are fed into the RoI-Pooling layer to get the features of proposals for the original and argument image: $z_{s}=RP(F_{s},P')$, $z_{a}=RP(F_{a},P')$.

After the pooling layer, the pooled feature $z_{s}$ and $z_{a}$ are fed into the downstream detection network, where the detection results are supervised by the loss function of Faster-RCNN. To facilitate domain alignment and promote feature diversity, $z_{s}$ and $z_{a}$ are sent into the Weighted Aligning Module (WAM) and Diversity Learning Module (DLM). The goal of this design is to achieve domain alignment while preserving domain-specific diversity. In the DLM, a feature diversity loss is introduced to enhance the variation of domain-specific features, ensuring robustness to unseen domains. Simultaneously, an entropy maximization loss is applied to estimate the category semantic and ensure the domain-specific information. In the WAM, dynamic loss weighting is employed to balance the contributions of domain alignment and diversity preservation. By integrating the supervision from both the DLM and WAM, the Diversity Invariance Detection Model (DIDM) achieves improved performance on the domain generalization task.

\subsection{\textbf{Diversity Learning Module (DLM)}}
\label{sec:methodDLM}
Existing generalization methods typically focus on learning domain-invariant features, which often leads to the suppression or loss of domain-specific information, resulting in suboptimal performance. To address this limitation, the Diversity Learning Module (DLM) is introduced. First, domain-specific features are extracted from the features by employing a two-branch network structure to compare and extract domain-specific characteristics. To fully leverage domain-specific patterns, an entropy maximization loss is applied to avoid category semantics in the domain-specific features. Second, a diversity loss is introduced to enhance the variability of domain-specific features. This loss explicitly promotes feature diversity, ensuring that domain-specific representations remain distinct and robust. By combining the entropy maximization loss and diversity loss, the DLM effectively enhances the diversity of domain-specific information, leading to improved generalization performance in domain generalization detection tasks.

Specifically, the Diversity Learning Module (DLM) operates as follows: first, to capture the variation of the augmented features relative to the original features, we draw inspiration from OCR \cite{jing2023order} and define the relationship between the original and augmented features as $z_{a}= \lambda_1 z_{s}+(1- \lambda_1) z_d$
where $\lambda_1$ is a gradually increasing hyperparameter, reflecting the model's improving capacity to learn domain-invariant features over the process of training. The difference between $z_{s}$ and $z_{a}$ diminishes as the model becomes less sensitive to domain-specific features. Ultimately, this approach enables us to obtain the domain-specific feature $z_d$ during training.

Second, considering that the domain-specific feature $z_d$ should not contain semantic information, we introduce the entropy maximization loss for $z_d$, where a classifier is trained with the cross-entropy loss:
\begin{equation}
\begin{aligned}
\mathcal{L}_\mathcal{C} = -\sum\limits_{i=1}^C y_ilog([Softmax(f(z_d))])
\end{aligned}
\end{equation}
where $C$ represents the number of categories, $f(z_d) \in R^{K \times C}$ is prediction function for $z_d$. Inspired by \cite{jing2023order}, to constrain the category semantic information in $z_d$, we implement to maximization of the entropy of the prediction results from $z_d$.
\begin{equation}
\begin{aligned}
\mathcal{L}_\mathcal{H} = -\mathcal{H}(y|z_d) = -\mathcal{H}[Softmax(f(z_d))]
\label{eq:pythagoras}
\end{aligned}
\end{equation}
where $\mathcal{H}$ represents entropy computation. By minimizing Eq. \ref{eq:pythagoras}, we can maximize the conditional entropy based on the classifier $f(.)$. When the output of the classifier for all categories is uniform across all categories, the conditional entropy reaches its maximum, thereby restricts the semantic information contained in $f(z_d)$.

Lastly, in addition to limiting semantic information, we introduce a \textbf{feature diversity (FD) loss} to promote diverse characteristics among features. Specifically, let the domain-specific feature for each proposal be $z_d^i$. We calculate the cosine similarity between each pair $z_d^i$ and $z_d^j$ for all proposals. Given that each proposal may capture different information due to variations in location or domain, our goal is to increase the distinctions between $z_d^i$ and $z_d^j$, where $i,j \in [1, K]$ and $K$ is the number of proposals. Consequently, the feature diversity loss can be expressed as:
\begin{equation}
\begin{aligned}
\mathcal{L}_\mathcal{FD} = -\sum\limits_{i=1}^K log(\dfrac{e^{(sim(f'(z_d^i),f'(z_d^i))/\tau)}}{\sum\limits_{j=1}^K e^{(sim(f'(z_d^i),f'(z_d^j))/\tau)}} )
\label{eq:pythagoras7}
\end{aligned}
\end{equation}
Where $\tau$ denotes the temperature hyperparameter,  which is utilized to control for differences between features and to encourage the model to achieve greater discrimination between various domain-specific features $z_d$. Here $sim(.)$ denotes the cosine similarity between feature mappings. $\hat{y}_i$ is set as 1 when $i=j$ and to 0 otherwise. $f'(.)$ is the function of the normalization layer. The diversity loss is designed to encourage a wide range of predicted categories, maximizing the dissimilarity among features.

The total loss for the Diversity Learning Module is calculated as follows:
\begin{equation}
\begin{aligned}
\mathcal{L}_\mathcal{DLM} = \mathcal{L}_\mathcal{C}+\mathcal{L}_\mathcal{H}+\lambda_2 \mathcal{L}_\mathcal{FD}
\end{aligned}
\end{equation}
Where $\lambda_2$ is the hyperparameter for balancing the loss. Under the supervision of the DLM, our model minimizes the semantic information in domain-specific features while maintaining their diversity, thereby enhancing the diversity of source domain distributions during training and improving the model’s robustness to unseen target domains.


\subsection{\textbf{Weighted Aligning Module (WAM)}}
\label{sec:methodWAM}
In addition to preserving feature diversity through the Diversity Learning Module (DLM), we propose the Weighted Alignment Module (WAM) to learn domain-invariant features while maintaining feature diversity. Traditional feature alignment methods focus solely on addressing domain discrepancy during domain alignment, which may lead to an over-tight domain-invariant feature space. This kind of distribution over-smoothing compromises feature diversity—a critical factor for generalization performance on the unseen target domain. Therefore, our goal is not to maximize alignment but to achieve a dynamic trade-off between diversity and alignment. To this end, the proposed WAM dynamically adjusts the alignment loss strength via a variable weight parameter.

Specifically, during the training phase, we align the original feature $z_{s}$ and the argument features $z_{a}$:
\begin{equation}
\begin{aligned}
\mathcal{L}(z_{s},z_{a}) =& -\dfrac{sim(z_{s}^i,z_{a}^i)}{sim(z_{s}^i,z_{a}^i) + \sum\limits_{j \neq i, j=1}^K sim(z_{s}^i,z_{a}^j)}
\end{aligned}
\end{equation}

By aligning features across different domains, we encourage the model to learn domain-invariant features. However, solely constraining domain disparity can reduce feature diversity. To achieve effective feature alignment while preserving feature diversity, we introduce a weight parameter that dynamically adjusts the weight of the feature alignment loss based on the similarity between the detection results of \( z_{s} \) and \( z_{a} \). Specifically, when detection results based on $z_{s}$ and $z_{a}$ are similar, the differences between $z_{s}$ and $z_{a}$ have minimal impact, suggesting that the alignment loss should be down-weighted to preserve feature diversity. The weight for the alignment loss is defined as:
\begin{align}
\beta &= 2-e^{-\dfrac {1}{2}(\mathcal{L}_{c}+\mathcal{L}_{b})}
\label{eq:pythagoras10}
\end{align}
Where $\mathcal{L}_{c} = \sum\limits_{i=1}^K KL(p_{s}^i,p_{a}^i)$, and
$\mathcal{L}_{b} = ||b_{s}^i-b_{a}^i||_2^2$ originates from \cite{danish2024improving}. $p_{s}^i, p_{a}^i$ are the classification results of $z_{s}$ and $z_{a}$, respectively, while $b_{s}^i-b_{a}^i$ are the bounding box regression results. Finally, the loss function of the weighted alignment module (WAM) is defined as follows:
\begin{equation}
\mathcal{L}_\mathcal{WAM}=\beta (1 + \mathcal L(z_s,z_a))
\end{equation}

By training with the weight parameters $\beta$, the model decreases the weight of the alignment loss when the distinctions between the original and augmented features do not influence the detection results. This dynamic adjustment mechanism enables the model to balance the relationship between feature alignment and feature diversity throughout the training process. Consequently, the WAM further enhances the model's ability to learn features.

\subsection{\textbf{Overall Optimization Objective}}
Overall optimization objective of the model is as follows:
\begin{align}
\mathcal{L}_{total}=\mathcal{L}_{det}+\alpha (\mathcal{L}_\mathcal{DLM}+ \mathcal{L}_\mathcal{WAM})
\label{eq:pythagoras12}
\end{align}
Here $\mathcal{L}_{det}=\mathcal{L}_{reg}+\mathcal{L}_{cls}$ is the detection loss, $\alpha$ is the hyperparameters for balancing the loss. With the co-training of the detection loss, DLM and WMA, our model achieve effective domain generalization on the task.


\begin{table*}
\caption{Single-Domain Generalization Object Detection Results. The Results Presented in the Table Indicate that Our Diversity Invariant Detection Model (DIDM) Significantly Enhances the Model's Generalization Performance, with Bold Text Highlighting Optimal Performance. This Underscores the Effectiveness of DIDM Across Various Environments.}
    \centering
    \setlength{\tabcolsep}{4.6pt}
    \fontsize{9}{15}\selectfont\begin{tabular}{c|c|c|cccc|c}
    \bottomrule
         \textbf{Methods}& \textbf{Ref}& \textbf{Daytime - Clear} &  \textbf{Night - Sunny }&  \textbf{Dusk - Rainy}&  \textbf{Night - Rainy}&  \textbf{Daytime - Foggy}&  \textbf{Average} \\
         \hline
          Faster R-CNN \cite{ren2016faster}& NeurIPS'15 & 57.3 &  38.1& 32.3 & 14.6 &  34.5& 29.9 \\
          SW \cite{pan2019switchable}& ICCV'19 &  50.6&  33.4&  26.3&  13.7&  30.8& 26.1\\
         IBN-Net \cite{pan2018two}& ECCV'18 &  49.7&  32.1&  26.1&  14.3&  29.6& 25.5\\
         IterNorm \cite{huang2019iterative}& CVPR'19 &  43.9&  29.6&  22.8&  12.6&  28.4& 23.4\\
         ISW \cite{choi2021robustnet}& CVPR'21 &  51.3&  33.2&  25.9&  14.1&  31.8& 26.3\\
         SHADE \cite{zhao2022style}& CVPR'22 &- &36.6 &29.5 &16.8 &33.9 &28.4\\
         S-DGOD \cite{wu2022single}& CVPR'22 &56.1 &36.6 &28.2 &16.6 &33.5 &28.8\\
         CLIP-Gap \cite{vidit2023clip}& CVPR'23 &51.3 &36.9 &32.3 &18.7 &38.5 &31.6 \\
         SRCD \cite{rao2024srcd}& TNNLS'24 &- &36.7 &28.8 &17.0 &35.9 &29.6\\
         Prompt-Driven \cite{li2024prompt}& CVPR'24 &53.6 &38.5 &33.7 &19.2 &39.1 &32.6 \\
         Diversification \cite{danish2024improving}& CVPR'24 &50.6 &39.4 &37.0 &22.0 &35.6 &33.5 \\
         DivAlign \cite{danish2024improving}& CVPR'24 &52.8 &42.5 &38.1 &24.1 &37.2 &\textbf{35.5} \\
         OA-DG \cite{lee2024object}& AAAI'24 &55.8 &38.0 &33.9 &16.8 &38.3 &31.8 \\
         UFR \cite{liu2024unbiased}& CVPR'24 &\textbf{58.6} &40.8 &33.2 &19.2 &39.6 &33.2 \\
         Sdg-yolov8 \cite{wang2025sdg} & Elsevier'25 &- &41.1 &39.0 
         &\textbf{24.6} &36.0 &35.2 \\
         DRSF \cite{li2025let} & arxiv'25 &57.7 &40.2 &\textbf{39.8} &20.2 &38.6 &34.7 \\
         TGFD\cite{wang2025enhanced} & Elsevier'25 &52.4 &36.8 &31.3 &16.7 &38.2 & 30.8 \\
         \hline
         \hline
         Ours& -  & 57.7 & \textbf{43.4} & 36.3 & 20.4 & \textbf{40.1} &35.1 \\
         \toprule
    \end{tabular}
    \label{tab:my_label1}
\end{table*}

\section{Experiments}
In this section, we conduct several experiments on the proposed DIDM model with different transfer tasks, including DWD datasets, Real to Artistic task, Diversified Weather task, and Artificial to Real task.

\label{sec:experiment}
\subsection{Experimental Setup}
\textbf{Implementation Details.} We utilize Faster R-CNN \cite{ren2016faster} as a base detector, with ResNet101 \cite{deng2021unbiased} backbone. The ImageNet \cite{deng2009imagenet} pretrained weights are employed for initialization. To train the model, we apply the stochastic gradient descent (SGD) algorithm with a momentum parameter of 0.9. The initial learning rate is set to 0.02 and decays every three epochs. All reported mean average precision (mAP) values are based on an Intersection over Union (IoU) threshold of 0.5.

\textbf{Data Augmentation Setting.} To further enhance the generalization of the model to the source domain data, a series of data augmentation techniques are employed to expand the source domain dataset. These techniques include random clipping, greyscale enhancement, and color transformation, which aim to increase the model's resilience to color variations. These augmentation methods significantly enrich the diversity of the dataset, allowing the model to be exposed to a wider range of features during the training process, thereby demonstrating improved adaptability and generalization capabilities in practical applications.

\subsection{Datasets} 
\textbf{DWD datasets} are included: Daytime - Clear, Night - Sunny, Dusk - Rainy, Night - Rainy, Daytime - Foggy. These datasets are primarily derived from 27,708 images captured during daytime-clear conditions and 26,158 images taken at night-sunny conditions, collected from BDD-100K \cite{yu2020bdd100k}. Additionally, 3,775 images with foggy conditions were sourced from Adverse Weather \cite{hassaballah2020vehicle} and Cityscape datasets \cite{cordts2016cityscapes}. The datasets with rain rendered using \cite{wu2021vector}, including 3,501 images from dusk-rainy days and 2,494 images from rainy nights. In this paper, we employ 19,395 daytime images of sunny conditions as the training set to train the model, while another 8,313 images are used to evaluate the model's performance. The remaining four datasets are treated as unseen target domains to assess the model's generalization capabilities. In these datasets, we focus on seven common categories: car, bike, bus, rider, person, motor, and truck.

\textbf{Real to Artistic} comprises four distinct types of datasets: Pascal VOC \cite{everingham2010pascal}, Clipart1k \cite{inoue2018cross}, Watercolor2k \cite{inoue2018cross}, and Comic2k \cite{inoue2018cross}. In this study, we utilize the Pascal VOC 2007 and VOC 2012 datasets, which consist of real images, to train the model. Subsequently, we then test the model on three additional datasets to evaluate its generalization performance on art-style images. Specifically, the Clipart1k dataset includes 20 categories that overlap with those in Pascal VOC, featuring a total of 1,000 clipart-style images. In contrast, Watercolor2k and Comic2k each contain 6 categories, representing subsets of the Pascal VOC categories, with Watercolor2k focusing on watercolor styles and Comic2k on manga styles.

\textbf{Diversified Weather} comprises datasets from four distinct environments: Cityscapes \cite{cordts2016cityscapes}, Foggy Cityscapes\cite{sakaridis2018semantic}, Rainy Cityscapes \cite{hu2019depth}, and BDD-100K \cite{yu2020bdd100k}. The Cityscapes datasets offer high-quality imagery of urban street scenes, while Foggy Cityscapes, Rainy Cityscapes, and BDD-100K further enhance the model's capability to comprehend various environments. In this study, we train the model using one of the datasets and subsequently evaluate its performance with the remaining datasets to assess its adaptability and robustness in unfamiliar environments.

\textbf{Artificial to Real} is a collection of synthetic images and two datasets from different environments: Sim10k \cite{2017Driving}, Cityscapes \cite{cordts2016cityscapes}, and BDD-100K \cite{yu2020bdd100k}. SIM10k \cite{2017Driving} consists of 10,000 artificial images rendered by CTA-V, which serve as the source domain for training the model. The model is then tested on the other datasets, demonstrating detection results in the car category to validate the feasibility of its cross-domain performance.


\begin{table}
\caption{The Quantitative Results (\%) on the Night-Sunny.}
    \centering
    \setlength{\tabcolsep}{4.1pt}
     \fontsize{7.6}{12.5}\selectfont\begin{tabular}{c|ccccccc|c}
    \bottomrule
     Methods & Bus & Bike & Car &Motor & Person & Rider &Truck & mAP \\
        \hline
           FR \cite{ren2016faster} & 39.8 & 31.1 & 64.0 & 17.6 & 43.0 & 28.3 & 42.7 & 38.1 \\
           SW \cite{pan2019switchable} &38.7 &29.2 &49.8 &16.6 &31.5 &28.0 &40.2 &33.4 \\
           IBN-Net \cite{pan2018two} &37.8 &27.3 &49.6 &15.1 &29.2 &27.1 &38.9 &32.1 \\
           IterNorm \cite{huang2019iterative} &38.5 &23.5 &38.9 &15.8 &26.6 &25.9 &38.1 &29.6 \\
           ISW \cite{choi2021robustnet} &38.5 &28.5 &49.6 &15.4 &31.9 &27.5 &41.3 &33.2 \\
         S-DGOD \cite{wu2022single} &40.6 &35.1 &50.7 &19.7 &34.7 &32.1 &43.4 &36.6\\
         CLIP-Gap \cite{vidit2023clip} &37.7 &34.3 &58.0 &19.2 &37.6 &28.5 &42.9 &36.9 \\
         SRCD \cite{rao2024srcd} &43.1 &32.5 &52.3 &20.1 &34.8 &31.5 &42.9 &36.7 \\
         Prompt-D \cite{li2024prompt}&\textbf{49.9} &35.0 &59.0 &21.3 &40.4 &29.9 &42.9 &38.5 \\
         UFR \cite{liu2024unbiased} &43.6 &38.1 &\textbf{66.1} &14.7 &\textbf{49.1} &26.4 &\textbf{47.5} &40.8 \\
        \hline
        \hline
         Ours&43.8  &\textbf{40.0}  &\textbf{66.1}  &\textbf{25.2}  &46.5  &\textbf{36.1}  &46.3  &\textbf{43.4} \\
         \toprule
    \end{tabular}
    \label{tab:my_label2}
\end{table}

\subsection{Performance on DWD datasets}
The DIDM we studied is compared with four feature-based normalization methods they are SW \cite{pan2019switchable}, IBN-Net \cite{pan2018two}, IterNorm \cite{huang2019iterative} and ISW \cite{choi2021robustnet}. Comparisons were also made with the latest SHADE \cite{zhao2022style}, S-DGOD \cite{wu2022single}, CLIP-Gap \cite{vidit2023clip}, SRCD \cite{rao2024srcd}, Prompt-Driven \cite{li2024prompt}, DivAlign \cite{danish2024improving}, OA-DG \cite{lee2024object}, UFR \cite{liu2024unbiased}, SDG \cite{wang2025sdg}, DRSF \cite{li2025let} and TGFD\cite{wang2025enhanced} methods.

\textbf{Results on DWD datasets.}
Table \ref{tab:my_label1} presents the testing and generalization results of the DIDM on the DWD datasets \cite{wu2022single}. In this study, we utilized the Daytime-Clear datasets to train the model, tested it on the same datasets, and evaluated its generalization capabilities on four additional datasets: Night-Sunny, Dusk-Rainy, Night-Rainy, and Daytime-Foggy. As shown in Table \ref{tab:my_label1}, our approach achieves an average generalization performance of 35.0\% and performs well in several scenarios, particularly under Night - Sunny and Daytime - Foggy conditions, where it attains the best performance metrics of 43.2\% and 40.1\%, respectively. When compared to the baseline network, Faster R-CNN \cite{ren2016faster}, our method demonstrates improvements of 5.6\% and 4.0\% on the Daytime-Foggy and Dusk-Rainy datasets, respectively. Additionally, performance on the Night-Sunny datasets was significantly enhanced by 5.1\%, while the Night-Rainy datasets, characterized by a complex environment, saw an improvement of 5.8\%. This demonstrates the robustness and adaptability of our method across different environments, particularly in challenging scenarios. The performance increase is caused by enhancing the feature distribution enhancement with feature diversity, where the distance between the augmented source domain and the target domain is reduced.


\begin{table}
\caption{The Quantitative Results (\%) on the Night - Rainy.}
    \centering
    \setlength{\tabcolsep}{4.1pt}
     \fontsize{7.6}{12.5}\selectfont\begin{tabular}{c|ccccccc|c}
    \bottomrule
     Methods & Bus & Bike & Car &Motor & Person & Rider &Truck & mAP \\
        \hline
           FR \cite{ren2016faster} & 23.9 & 7.1 & 36.5 & 0.2 & 9.9 & 7.5 & 17.0 & 14.6 \\
           SW \cite{pan2019switchable} &22.3 &7.8 &27.6 &0.2 &10.3 &10.0 &17.7 &13.7 \\
           IBN-Net \cite{pan2018two} &24.6 &10.0 &28.4 &0.9 &8.3 &9.8 &18.1 &14.3 \\
           IterNorm \cite{huang2019iterative} &21.4 &6.7 &22.0 &0.9 &9.1 &10.6 &17.6 &12.6 \\
           ISW \cite{choi2021robustnet} &22.5 &11.4 &26.9 &0.4 &9.9 &9.8 &17.5 &14.1 \\
           S-DGOD \cite{wu2022single} &24.4 &11.6 &29.5 &9.8 &10.5 &11.4 &19.2 &16.6\\
          CLIP-Gap \cite{vidit2023clip} &28.6 &12.1 &36.1 &9.2 &12.3 &9.6 &22.9 &18.7 \\
          SRCD \cite{rao2024srcd} &26.5 &12.9 &32.4 &0.8 &10.2 &12.5 &24.0 &17.0 \\
          Prompt-D \cite{li2024prompt}&25.6 &12.1 &35.8 &\textbf{10.1} &\textbf{14.2} &\textbf{12.9} &22.9 &19.2 \\
          UFR \cite{liu2024unbiased} &29.9 &11.8 &36.1 &9.4 &13.1 &10.5 &23.3 &19.2 \\
        \hline
        \hline
         Ours&\textbf{33.8}  &\textbf{19.2}  &\textbf{41.2}  &1.2  &11.6  &11.2 &\textbf{24.7}  &\textbf{20.4} \\
         \toprule
    \end{tabular}
    \label{tab:my_label4}
\end{table}
\begin{table}
\caption{The Quantitative Results (\%) on the Daytime - Clear.}
    \centering
    \setlength{\tabcolsep}{4.1pt}
     \fontsize{7.6}{12.5}\selectfont\begin{tabular}{c|ccccccc|c}
    \bottomrule
     Methods & Bus & Bike & Car &Motor & Person & Rider &Truck & mAP \\
        \hline
           FR \cite{ren2016faster} & 59.5 & 47.6 & 76.7 & 49.3 & 56.0 & 51.4 & 60.3 & 57.3 \\
           SW \cite{pan2019switchable} &62.3 &42.9 &53.3 &49.9 &39.2 &46.2 &60.6 &50.6 \\
           IBN-Net \cite{pan2018two} &63.6 &40.7 &53.2 &45.9 &38.6 &45.3 &60.7 &49.7 \\
           IterNorm \cite{huang2019iterative} &58.4 &34.2 &42.4 &44.1 &31.6 &40.8 &55.5 &43.9 \\
           ISW \cite{choi2021robustnet} &62.9 &44.6 &53.5 &49.2 &39.9 &48.3 &60.9 &51.3 \\
         S-DGOD \cite{wu2022single} &\textbf{68.8} &50.9 &53.9 &56.2 &41.8 &52.4 &\textbf{68.7} &56.1\\
         CLIP-Gap \cite{vidit2023clip} &55.0 &47.8 &67.5 &46.7 &49.4 &46.7 &54.7 &52.5 \\
         Prompt-D \cite{li2024prompt}&\- &- &- &- &- &- &- &53.6 \\
         UFR \cite{liu2024unbiased} &66.8 &\textbf{51.0} &70.6 &\textbf{55.8} &49.8 &48.5 &67.4 &\textbf{58.6} \\
        \hline
        \hline
         Ours&60.5  &48.6  &\textbf{76.3}  &49.3  &\textbf{56.0}  &\textbf{53.2}  &60.4  &57.7 \\
         \toprule
    \end{tabular}
    \label{tab:my_label_dayclear}
\end{table}

\textbf{Results on Night-Sunny conditions.}
As shown in the results presented in Table \ref{tab:my_label2}, our method exhibits outstanding performance across several object detection categories, particularly in the Bus, Car, and bike categories, our method attains detection accuracies of 43.8\%, 66.1\%, and 40.0\%, respectively. This demonstrates the advantages of our approach in feature normalization and target detection. In addition, we achieve an overall performance improvement of 6.5\%, 6.7\%, 4.9\%, and 2.6\% compared to the latest CLIP-Gap \cite{vidit2023clip}, SRCD \cite{rao2024srcd}, prompt-D \cite{li2024prompt} and UFR \cite{liu2024unbiased}, respectively. This further demonstrates the superior adaptability and robustness of our method in handling data across different domains.

\begin{figure*}
    \centering
    \includegraphics[width=1\linewidth]{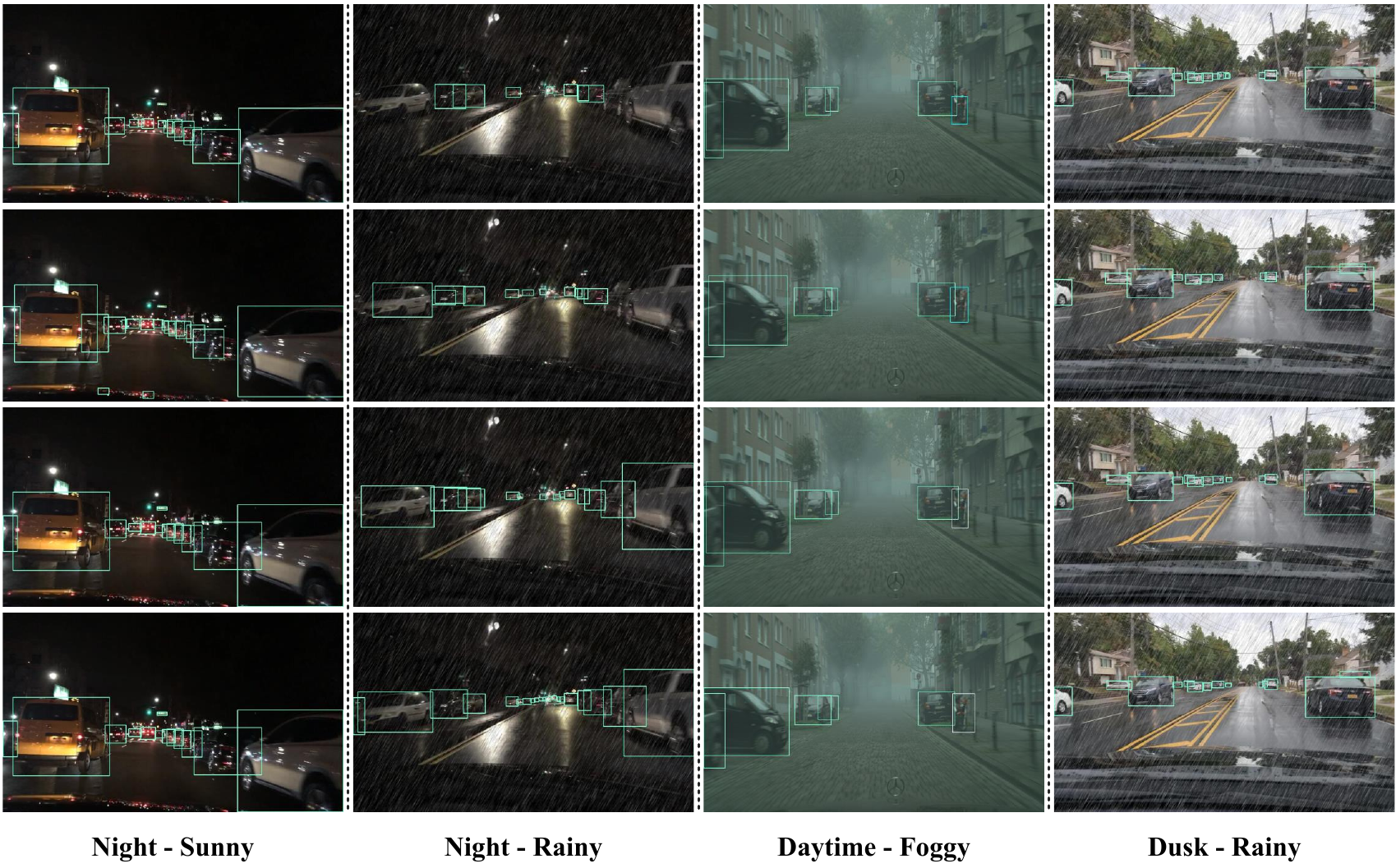}
    \caption{Qualitative evaluation results for night-sunny, night-rainy daytime-foggy and dusk-rainy weather conditions are presented. The images, arranged from top to bottom, display the visualization results generated by \textbf{Faster R-CNN} \cite{ren2016faster}, \textbf{CLIP-Gap} \cite{vidit2023clip}, \textbf{DIDM}, and the \textbf{Ground Truth}, respectively.}
    \label{fig:DWD}
\end{figure*}

\begin{table}
\caption{The Quantitative Results (\%) on the Daytime - Foggy.}
    \centering
    \setlength{\tabcolsep}{4.1pt}
     \fontsize{7.6}{12.5}\selectfont\begin{tabular}{c|ccccccc|c}
    \bottomrule
     Methods & Bus & Bike & Car &Motor & Person & Rider &Truck & mAP \\
        \hline
           FR \cite{ren2016faster} & 32.0 & 29.1 & 56.7 & 28.7 & 34.1 & 38.8 & 22.1 & 34.5 \\
           SW \cite{pan2019switchable} &30.6 &36.2 &44.6 &25.1 &30.7 &34.6 &23.6 &30.8 \\
           IBN-Net \cite{pan2018two} &29.9 &26.1 &44.5 &24.4 &26.2 &33.5 &22.4 &29.6 \\
           IterNorm \cite{huang2019iterative} &29.7 &21.8 &42.4 &24.4 &26.0 &33.3 &21.6 &28.4 \\
           ISW \cite{choi2021robustnet} &29.5 &26.4 &49.2 &27.9 &30.7 &34.8 &24.0 &31.8 \\
           S-DGOD \cite{wu2022single} &32.9 &28.0 &48.8 &29.8 &32.5 &38.2 &24.1 &33.5\\
          CLIP-Gap \cite{vidit2023clip} &36.1 &34.3 &58.0 &33.1 &39.0 &43.9 &25.1 &38.5 \\
          SRCD \cite{rao2024srcd} &36.4 &30.1 &52.4 &31.3 &33.4 &40.1 &27.7 &35.9 \\
          Prompt-D \cite{li2024prompt}&36.1 &\textbf{34.5} &58.4 &33.3 &\textbf{40.5} &44.2 &26.2 &39.1 \\
          UFR \cite{liu2024unbiased} &36.9 &35.8 &61.7 &\textbf{33.7} &39.5 &42.2 &27.5 &39.6 \\
        \hline
        \hline
         Ours&\textbf{41.1}  &33.2  &\textbf{62.5}  &33.5  &37.2  &\textbf{44.7}  &\textbf{28.7}  &\textbf{40.1} \\
         \toprule
    \end{tabular}
    \label{tab:my_label5}
\end{table}

\textbf{Results on Night-Rainy conditions.}
The Night - Rainy presents a particularly challenging situation. Low visibility and complex lighting significantly impacted the color-dependent model, while the rain further diminished visual clarity. Table \ref{tab:my_label4} illustrates that our method significantly outperforms other approaches in automobile category detection, particularly in bus detection, with improvements of 5.2\%, 7.3\%, 8.2\% and 3.9\% compared to CLIP-Gap \cite{vidit2023clip}, SRCD \cite{rao2024srcd}, prompt-D \cite{li2024prompt} and UFR \cite{liu2024unbiased}, respectively.

\begin{table}
\caption{The Quantitative Results (\%) on the Dusk - Rainy.}
    \centering
    \setlength{\tabcolsep}{4.1pt}
     \fontsize{7.6}{12.5}\selectfont\begin{tabular}{c|ccccccc|c}
    \bottomrule
     Methods & Bus & Bike & Car &Motor & Person & Rider &Truck & mAP \\
        \hline
           FR \cite{ren2016faster} & 38.7 & 25.8 & 66.3 & 9.7 & 28.3 & 17.1 & 40.6 & 32.3 \\
           SW \cite{pan2019switchable} &35.2 &16.7 &50.1 &10.4 &20.1 &13.0 &38.8 &26.3 \\
           IBN-Net \cite{pan2018two} &37.0 &14.8 &50.3 &11.4 &17.3 &13.3 &38.4 &26.1 \\
           IterNorm \cite{huang2019iterative} &32.9 &14.1 &38.9 &11.0 &15.5 &11.6 &35.7 &22.8 \\
           ISW \cite{choi2021robustnet} &34.7 &16.0 &50.0 &11.1 &17.8 &12.6 &38.8 &25.9 \\
         S-DGOD \cite{wu2022single} &37.1 &19.6 &50.9 &13.4 &19.7 &16.3 &40.7 &28.2\\
         CLIP-Gap \cite{vidit2023clip} &37.8 &22.8 &60.7 &16.8 &26.8 &18.7 &42.4 &32.3 \\
         SRCD \cite{rao2024srcd} &39.5 &21.4 &50.6 &11.9 &20.1 &17.6 &40.5 &28.8 \\
         Prompt-D \cite{li2024prompt}&39.4 &25.2 &60.9 &\textbf{20.4} &\textbf{29.9} &16.5 &43.9 &33.7 \\
         UFR \cite{liu2024unbiased} &37.1 &21.8 &67.9 &16.4 &27.4 &17.9 &43.9 &33.2 \\
        \hline
        \hline
         Ours&\textbf{41.9}  &\textbf{30.1}  &\textbf{68.2}  &18.8  &\textbf{29.9}  &\textbf{19.9}  &\textbf{45.2}  &\textbf{36.3} \\
         \toprule
    \end{tabular}
    \label{tab:my_label3}
\end{table}

\textbf{Results on Daytime-Clear conditions.}
We evaluate the proposed method in this paper using a dataset that belongs to the same domain as the training set, as illustrated in Table \ref{tab:my_label_dayclear}. The results indicate that the proposed method performs exceptionally well across several detection categories, particularly in the Car, Person, and Rider categories. It shows a distinct advantage over other methods, achieving of 76.3\%, 56.0\%, and 53.2\%, respectively. Compared to the baseline Faster R-CNN \cite{ren2016faster} method, this approach demonstrates notable improvements, thereby validating its effectiveness in the target detection task.

\textbf{Results on Daytime-Foggy conditions.}
As shown in Table \ref{tab:my_label5}, our method demonstrates superior generalization performance under daytime-foggy conditions, achieving 40.1\% of the best detection performance compared to all methods listed in the table.

\textbf{Results on Dusk-Rainy conditions.}
As shown in Table \ref{tab:my_label3}, our method demonstrates exceptional performance across all categories, particularly in the Bus, Car, and Truck categories. Furthermore, the overall performance improves by 2.6\% and 3.1\% compared to the Prompt-D \cite{li2024prompt} and UFR \cite{liu2024unbiased}, the most recent methods, respectively.

\begin{figure*}
    \centering
    \includegraphics[width=1\linewidth]{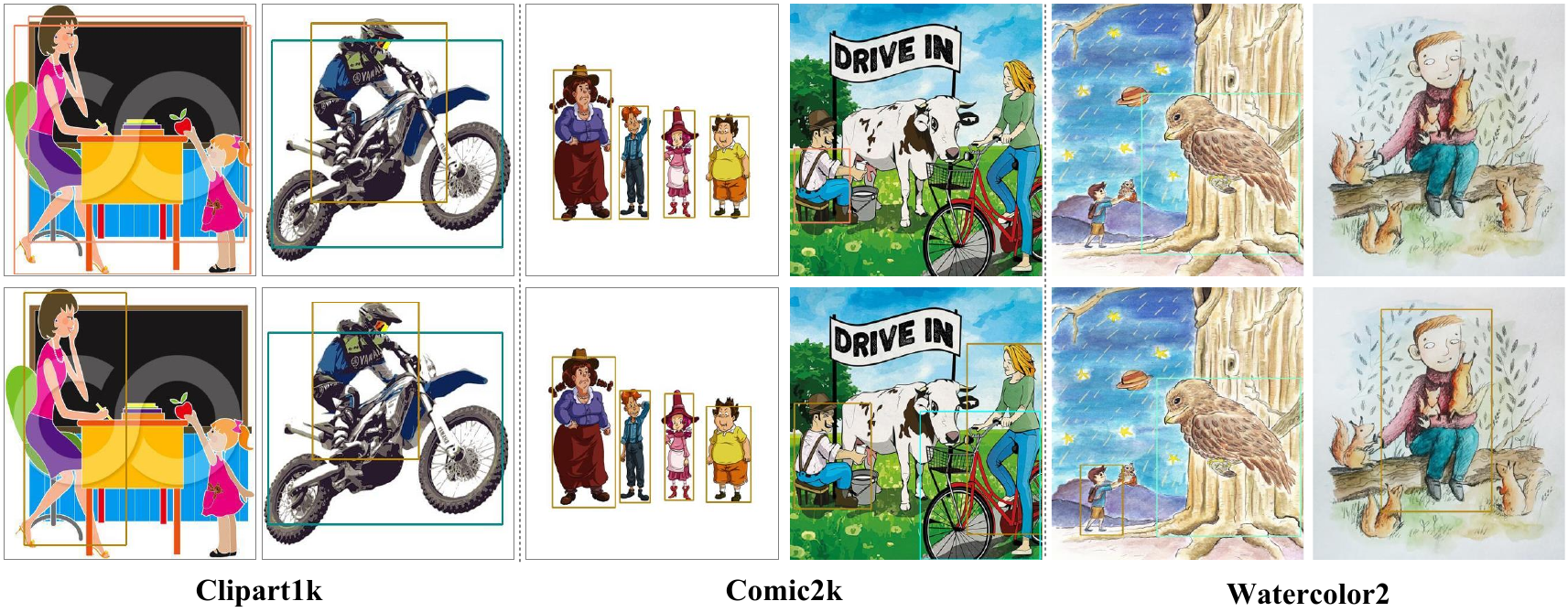}
    \caption{Qualitative evaluation results for Real to Artistic datasets are presented. The \textbf{top row} showcases the detection visualizations generated by the \textbf{Faster R-CNN} \cite{ren2016faster} model, while the \textbf{bottom row} displays the detection visualizations produced by the \textbf{DIDM}.}
    \label{fig:VOC}
\end{figure*}
\begin{table}
\caption{DA and DG Results on Real to Artistic Datasets}
    \centering
    \setlength{\tabcolsep}{11pt}
     \fontsize{9.5}{15}\selectfont\begin{tabular}{c|ccc}
    \bottomrule
        Methods & Clipart & Comic & watercolor\\
        \hline
        \multicolumn{4}{c}{DA methods} \\
        SWDA \cite{saito2019strong} &38.1  &27.4  &53.3 \\
        I3Net \cite{chen2021i3net}  & - &\textbf{30.1} & 51.5 \\
        HTCN \cite{chen2020harmonizing}  & 40.3 & - & -\\
        DBGL \cite{chen2021dual} & \textbf{41.6}  & 29.7 & \textbf{53.8} \\
        \hline
        \hline
        \multicolumn{4}{c}{DG methods} \\
        FR \cite{ren2016faster}  &25.1  &16.8  &45.5 \\
        NP \cite{fan2023towards}  &35.4  &28.9  &53.3 \\
        Div. \cite{danish2024improving}  & 33.7   & 25.5 & 52.5\\
        DivAlign \cite{danish2024improving}  & 38.9  & \textbf{33.2} & \textbf{57.4}\\
        GDD(R101) \cite{he2025generalized}  & \textbf{40.8}  & 29.7 & 54.2\\
        FGT(R101) \cite{he2025boosting}  & 40.5  & 30.0 & 56.6\\
        \hline
        Ours  &34.5  &31.7  &57.3 \\
        \toprule
    \end{tabular}
    \label{tab:my_label_voc}
\end{table}
\textbf{Visualization Results}
The visual analysis presented in Fig. \ref{fig:DWD} clearly demonstrates that balancing domain diversity and domain invariance is crucial for enhancing the model's generalization capabilities. The quantitative results indicate that the proposed method significantly outperforms the Faster R-CNN \cite{ren2016faster} baseline in detection visualization, effectively addressing the issues of misdetection and omission identified in CLIP-Gap \cite{vidit2023clip}, with its detection accuracy nearing the level of actual labeling. Notably, the DIDM method consistently exhibits exceptional recognition accuracy in challenging and complex scenarios, such as Night-Rainy conditions, thereby validating its effectiveness in improving the robustness of target detection.

\subsection{Real to Artistic task}
In this study, we thoroughly evaluate the generalization performance of the DIDM method for art style images. We provide a detailed comparison with the current mainstream domain generalization methods DivAlign \cite{danish2024improving}, NP \cite{fan2023towards}, GDD(R101) \cite{he2025generalized}, and FGT(R101) \cite{he2025boosting} as well as various domain adaptation methods SWDA \cite{saito2019strong}, I3Net \cite{chen2021i3net}, HTCN \cite{chen2020harmonizing} and DBGL \cite{chen2021dual}. The results are shown in Table \ref{tab:my_label_voc}. Within the three major domains—Clipart, Comic, and Watercolor, which exhibit significant stylistic differences, the method presented achieves stable performance, effectively validating the model's strong adaptability to complex distributional biases. Specifically, among the domain generalization (DG) methods, our approach achieves a detection accuracy of 57.3\% in the Watercolor domain, which is nearly equivalent to the current optimal DivAlign method (57.4\%), while significantly outperforming other DG benchmark methods. Our methods also outperforms some of DA models which require target domain data. Therefore, the effectiveness of the proposed method for enhancing the feature diversity is proven.

\textbf{Visualization Results} We also present the visualization result of the experiment in Fig. \ref{fig:VOC}, where the first and second rows show the detection results without transfer learning and our methods, respectively. Obviously, our method achieves better detection performance.


\begin{table*}
\caption{Single-Domain Generalization Object Detection Results. We trained the Model Using One dataset as the Source, while Also Assessing Its Generalization Across the Other Three Domains.}
    \centering
    \setlength{\tabcolsep}{9.5pt}
    \fontsize{9}{15}\selectfont\begin{tabular}{c|c|c|cccc}
    \bottomrule
         \textbf{\diagbox{Source}{Target}}& \textbf{Methods}& \textbf{Ref}& \textbf{Cityscaps} &  \textbf{Foggy Cityscapes}&  \textbf{Rainy Cityscapes} &\textbf{BDD-100K}  \\
         \hline

         \hline
         \multirow{5}{*}{Foggy Cityscapes \cite{sakaridis2018semantic}}
         & FR \cite{ren2016faster}& NeurIPS'15 &33.3 &- &43.7  &13.9  \\
         & FACT \cite{xu2021fourier} &CVPR'21 &30.0 &- &38.7  &20.2\\
         & FSDR \cite{huangfrequency} &CVPR'22 &31.3 &- &40.8   &20.4\\
         & MAD \cite{xu2023multi} &CVPR'23 &41.3 &- &43.3  &\textbf{24.4}\\
         & Ours &- &\textbf{44.5} &- &\textbf{50.2}  &24.1\\

         \hline
         \multirow{5}{*}{BDD-100K \cite{yu2020bdd100k}}
         & FR \cite{ren2016faster}& NeurIPS'15 &37.4 &26.9 &45.2  &-  \\
         & FACT \cite{xu2021fourier} &CVPR'21 &32.4 &24.3 &33.9  &-  \\
         & FSDR \cite{huangfrequency} &CVPR'22 &32.4 &27.8 &34.7   &-  \\
         & MAD \cite{xu2023multi} &CVPR'23 &36.4 &30.3 &36.1  &-   \\
         & Ours &-  &\textbf{38.2} &\textbf{31.4}  &\textbf{45.4}  &-    \\
         
         \toprule
    \end{tabular}
    \label{tab:Adverse}
\end{table*}

\begin{table}
\caption{The Quantitative Results (\%) from Cityscapes to BDD-100K.}
    \centering
    \setlength{\tabcolsep}{4.1pt}
     \fontsize{7.6}{12.5}\selectfont\begin{tabular}{c|ccccccc|c}
    \bottomrule
         Methods  & Bus & Bike & Car &Motor & Person & Rider  &Truck & mAP \\
         \hline
         FR \cite{ren2016faster}  &33.5  &26.8  &41.3  &13.0  &16.2  &11.2  &23.6   &23.7 \\
         SW \cite{pan2019switchable}   &22.2  &21.3  &36.7  &\textbf{20.9}  &25.6  &23.1  &18.8   &24.1 \\
         ISW \cite{choi2021robustnet}   &21.6  &20.9  &35.2  &17.8  &22.7  &22.4  &16.6   &22.5 \\
         IBN-Net \cite{pan2018two}   &19.2  &14.9  &31.9  &13.7  &21.4  &19.3  &13.4   &19.1 \\
         IterNorm \cite{huang2019iterative}   &21.6  &21.2  &36.2  &\textbf{20.9}  &23.4  &10.6  &18.0   &23.7 \\
         CDSD \cite{wu2022single}   &20.5  &22.9  &33.8  &14.7  &18.5  &23.6  &18.2   &21.7 \\
         SHADE \cite{zhao2022style}  &19.0  &25.1  &36.8  &18.4  &24.1  &24.9  &19.8   &24.0 \\
         SRCD \cite{rao2024srcd}   &21.5  &24.8  &38.7  &19.0  &\textbf{25.7}  &\textbf{28.4}  &23.1   &25.9 \\
         MAD \cite{xu2023multi}   &-  &-  &-  &-  &-  &-  &-   &28.0 \\
         \hline
         \hline
         Ours   &\textbf{40.4}  &\textbf{33.2}  &\textbf{56.0}  &19.2  &20.3  &19.4  &\textbf{30.4}  &\textbf{31.3} \\
         \toprule
    \end{tabular}
    \label{tab:bdd100k}
\end{table}

\subsection{Diversified Weather datasets}
Here, we compare DIDM with recent methods object detection such as FACT \cite{xu2021fourier}, FSDR \cite{huangfrequency}, and MAD \cite{xu2023multi}, as well as with four feature-based normalization techniques: SW \cite{pan2019switchable}, IBN-Net \cite{pan2018two}, IterNorm \cite{huang2019iterative}, and ISW \cite{choi2021robustnet}. This comparison aims to further validate the applicability and effectiveness of DIDM across various domains.

\textbf{Results on Diversified Weather.} In this study, we train the model using the only one datasets, and evaluate its performance on the remaining datasets to thoroughly assess the generalization capability of the DIDM method. As shown in Table \ref{tab:Adverse}, our approach exhibits outstanding performance across datasets with varying weather conditions. This highlights the effectiveness of balancing the diversity of domain-invariant features with domain-specific features during the learning process.

\textbf{Results on BDD-100K.} Table \ref{tab:bdd100k} presents a performance comparison of the method proposed in this paper with several contemporary DGOD methods across various categories. The results indicate that the DIDM method demonstrates exceptional generalization performance in all categories, particularly in the bus, car, and truck categories, where it significantly surpasses the other methods. This is a strong indication that the approach outlined in this paper achieves superior generalization and accuracy in the target detection task.

\begin{table}
\caption{Car Class Detection Results (\%) from Sim10k Dataset to Cityscapes and BDD100K Datasets.}
    \centering
    \setlength{\tabcolsep}{12pt}
     \fontsize{8}{12.5}\selectfont\begin{tabular}{c|c|cc}
    \bottomrule
        \textbf{Methods} & \textbf{Ref}  & \textbf{Cityscapes} & \textbf{BDD-100K} \\
        \hline
        FR \cite{ren2016faster} & NeurIPS'15  &34.0  &30.0 \\
        IBN-Net \cite{pan2018two}   & ECCV'18 &  33.2 & 25.7  \\
        SW \cite{pan2019switchable}   & ICCV'19  &  34.5 & 30.0  \\
        IterNorm \cite{huang2019iterative}  & CVPR'19 &  34.3 & 25.7  \\
        ISW \cite{choi2021robustnet}& CVPR'21 &  40.4 & 28.5  \\ 
        CDSD \cite{wu2022single}  & CVPR’22  &35.2  &27.4  \\
        SHADE \cite{zhao2022style}  &CVPR’22   & 40.9   & 30.3 \\
        SRCD \cite{danish2024improving}  &TNNLS’24   & 43.0  & 31.6 \\
        \hline
        \hline
        Ours  & -  &\textbf{51.6}  &\textbf{39.8}  \\
        \toprule
    \end{tabular}
    \label{tab:Artificial}
\end{table}

\subsection{Artificial to Real datasets}
Here, we compare DIDM with recent ISW \cite{choi2021robustnet}, CDSD \cite{wu2022single}, SRCD \cite{danish2024improving} and four feature-based normalization methods. As shown in Table \ref{tab:Artificial}, the results of car category detection on artificial-to-real datasets demonstrate that the proposed method exhibits exceptional performance on both the Cityscapes and BDD-100K datasets. It significantly outperforms the previous best SRCD \cite{danish2024improving} method, which achieved detection results of 43.0\% and 31.6\%, with our method achieving 51.6\% and 39.8\%, respectively. This fully validates its efficient migration capability and robustness in synthetic-to-real scenarios within the cross-domain detection task. 

\begin{table*}
\caption{The Results of Our Proposed Diversity Invariant Detection Model (DIDM) Ablation Experiment are Analyzed. The Losses Introduced in this Paper are Progressively Incorporated into the Baseline Network Faster R-CNN \cite{ren2016faster}, and the Contribution of Each Module to Target Detection is Examined.}
    \centering
    \setlength{\tabcolsep}{5.7pt}
    \fontsize{9.4}{16}\selectfont\begin{tabular}{c|ccc|c|cccc}
    \bottomrule
         Methods& $\mathcal L_\mathcal C+\mathcal L_\mathcal H$&  $\mathcal L_\mathcal{FD}$&  $\mathcal L_\mathcal{WAM}$&  \textbf{Daytime - Clear} &  \textbf{Night - Sunny }&  \textbf{Dusk - Rainy}&  \textbf{Night - Rainy}&  \textbf{Daytime - Foggy}\\
         \hline
         \multirow{6}{*}{FR \cite{ren2016faster}}
         &  &  & & 57.3 &  38.1 &  32.3 &  14.6 &  34.5 \\
         &  $\checkmark$&  &  &56.9  &41.9  &34.4 &17.1  &38.6 \\
         &  &$\checkmark$  &  & \textbf{58.0} & 41.8 & 33.3 & 16.8 &38.4 \\
         &  $\checkmark$&  $\checkmark$&  &57.7  &42.4  &35.7  &19.2  &39.4  \\
         &  &  &  $\checkmark$  &57.9  &42.2  &35.5 &18.7  &39.4\\
         &  $\checkmark$&  $\checkmark$&  $\checkmark$&  57.7&  \textbf{43.4}& \textbf{36.3}  &  \textbf{20.4}& \textbf{40.1}\\
    \toprule
    \end{tabular}
    \label{tab:my_label6}
\end{table*}

\section{Model Analysis and Discussion}
In this section, we analyze and discuss our model with several different experiment to show its characteristics and effectiveness. An in-depth insight into our model is presented.

\subsection{Ablation Study}
To assess the influence of each component on the performance of DIDM, we performed an ablation study. The model was trained using the daytime-clear datasets and subsequently evaluated on the remaining four datasets: night-sunny, dusk-rainy, night-rainy, and daytime-foggy. Through a comprehensive analysis of the individual contributions of DLM and WAM, we confirmed the efficacy of both modules in the context of object detection. Furthermore, we illustrated the synergistic effect achieved by integrating these two modules, which significantly improves the model's generalization capabilities and detection accuracy. The conclusive results are detailed in Table \ref{tab:my_label6}.
\begin{figure*}[!t]
\centering
\captionsetup[subfloat]{font=scriptsize}
\subfloat[]{\includegraphics[width=2.22in]{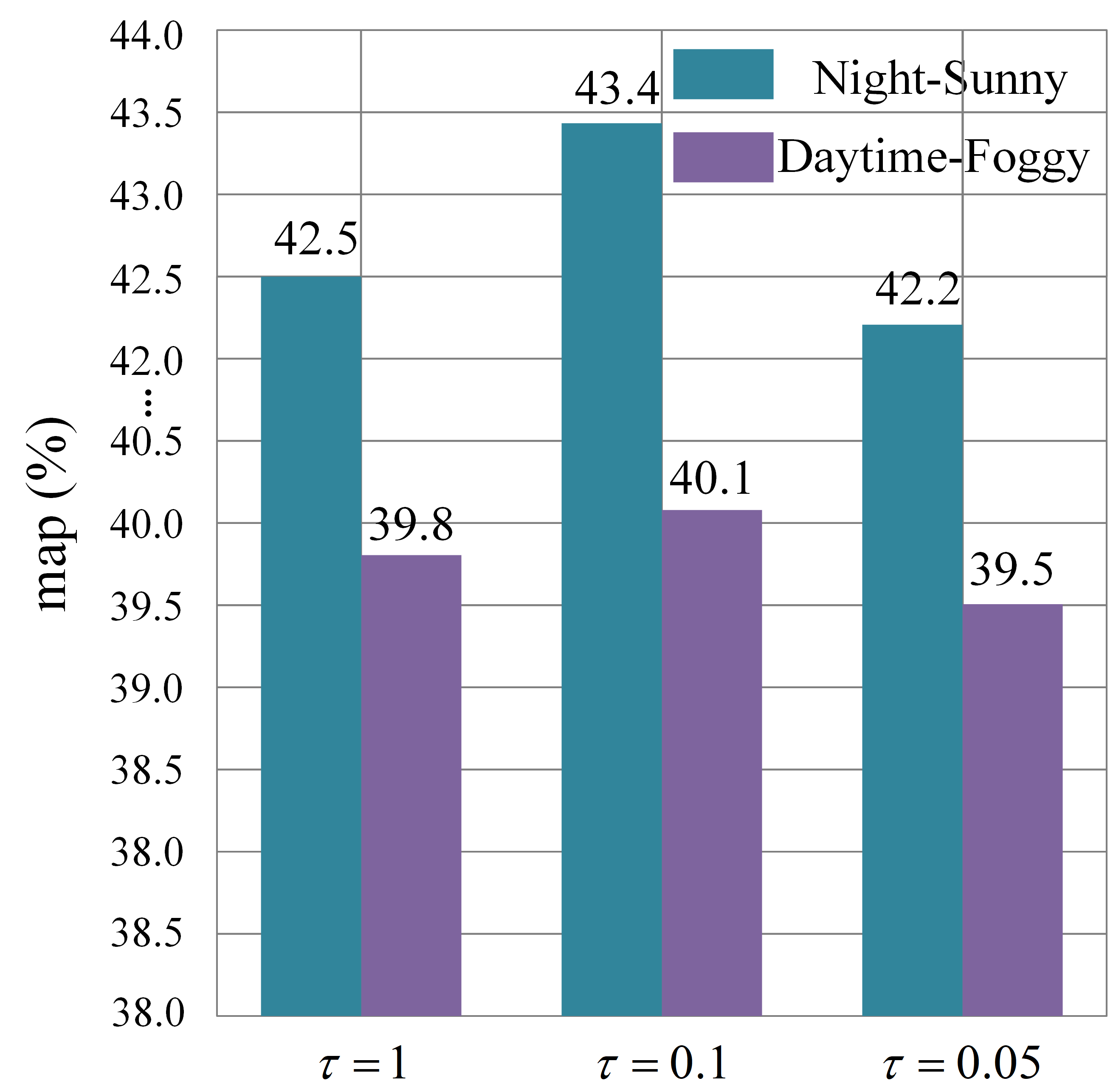}%
\label{fig:short-b}}
\subfloat[]{\includegraphics[width=1.62in]{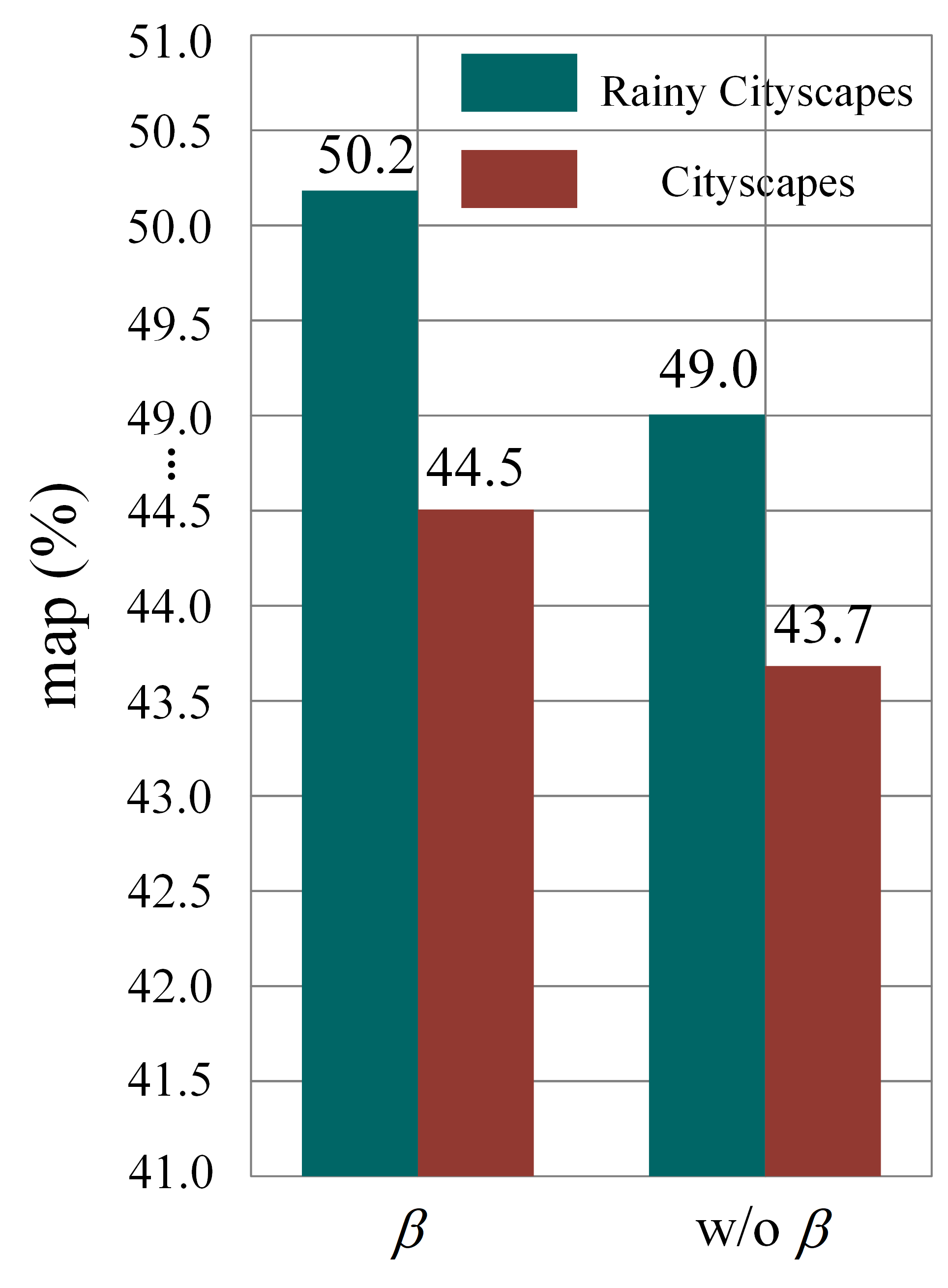}%
\label{fig:short-c}}
\subfloat[]{\includegraphics[width=3.2in]{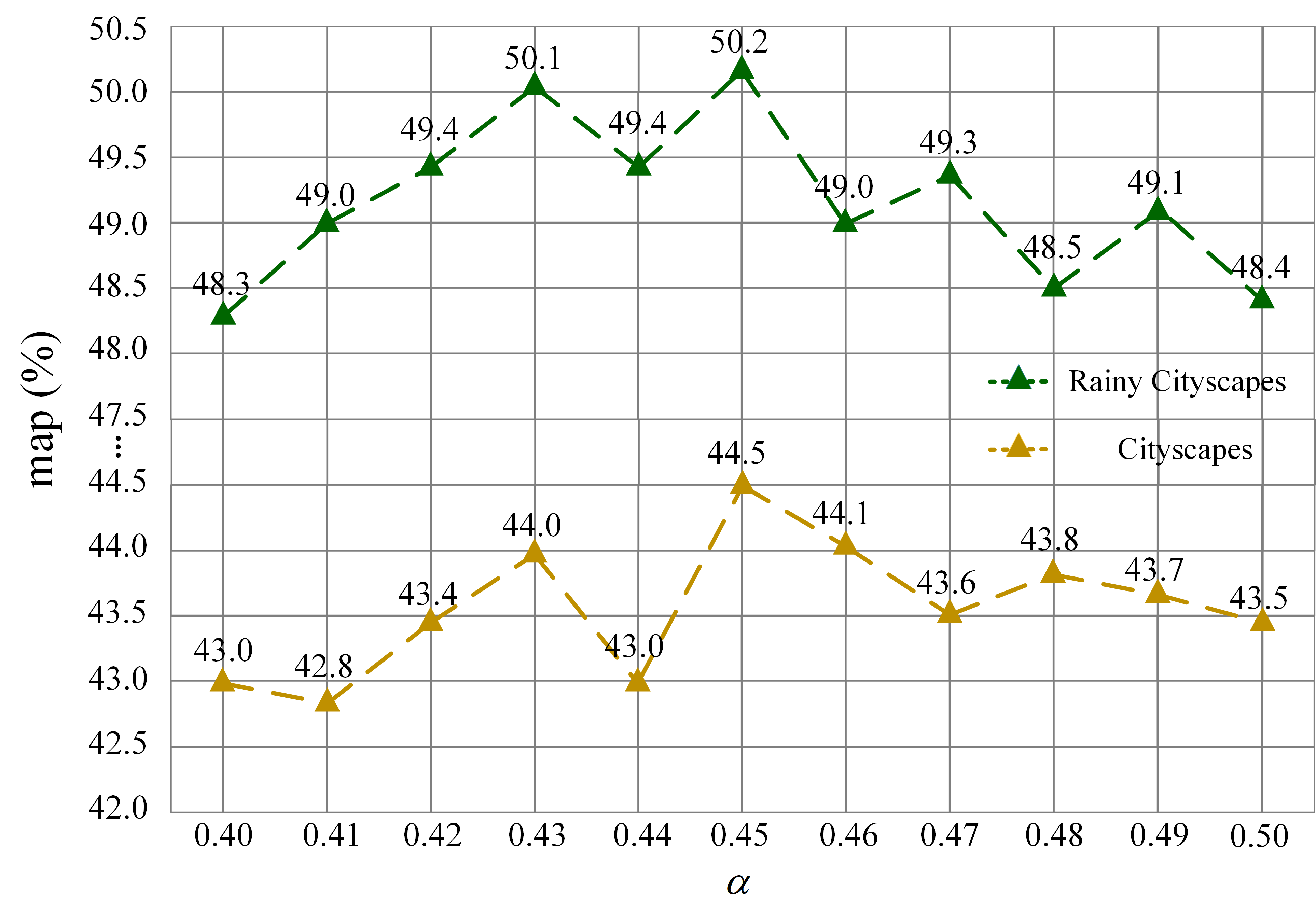}%
\label{fig:short-a}}
\caption{Hyperparameter Analysis. (a) Analysis of $\tau$. (b) Analysis of $\beta$. (c) Analysis of the Choice of Hyperparameter $\alpha$.}
\label{figs_sim}
\end{figure*}

The first component is the DLM, as illustrated in Table \ref{tab:my_label6}. The model's performance can be significantly improved by utilizing $\mathcal L_\mathcal H$ and $\mathcal L_\mathcal{FD}$, respectively. This suggests that either constraining the semantic information of domain-specific features or preserving the diversity of these features can enhance the model's detection capabilities. Furthermore, the model's performance is further augmented by employing both $\mathcal L_\mathcal H$ and $\mathcal L_\mathcal{FD}$. This demonstrates that the DLM component effectively enhances the model's ability to extract semantic information from images. In  Table \ref{tab:my_label6}, we analyze the role of the WAM and observe that the model's generalization performance improves by 4.9\% and 4.3\% for daytime-foggy and night-sunny conditions, respectively, compared to the baseline Faster R-CNN \cite{ren2016faster}. Additionally, the model's test MAP achieves a score of 57.9\%. This indicates that the WAM enhances the model's ability to learn domain-invariant features while maintaining a degree of diversity in domain-specific features by incorporating loss weights.

Ultimately, we integrated the DLM and WAM components and found that the model's generalization performance was significantly improved while maintaining its testing performance to a certain extent. This indicates that the synergy between DLM and WAM effectively balance between the diversity of domain-specific and invariance cross domains. This balance enhances the model's learning and feature extraction capabilities while ensuring stable detection performance across various environments.

\begin{figure*}[!t]
\captionsetup[subfloat]{font=scriptsize}
\centering
\subfloat[]{\includegraphics[width=2.3in]{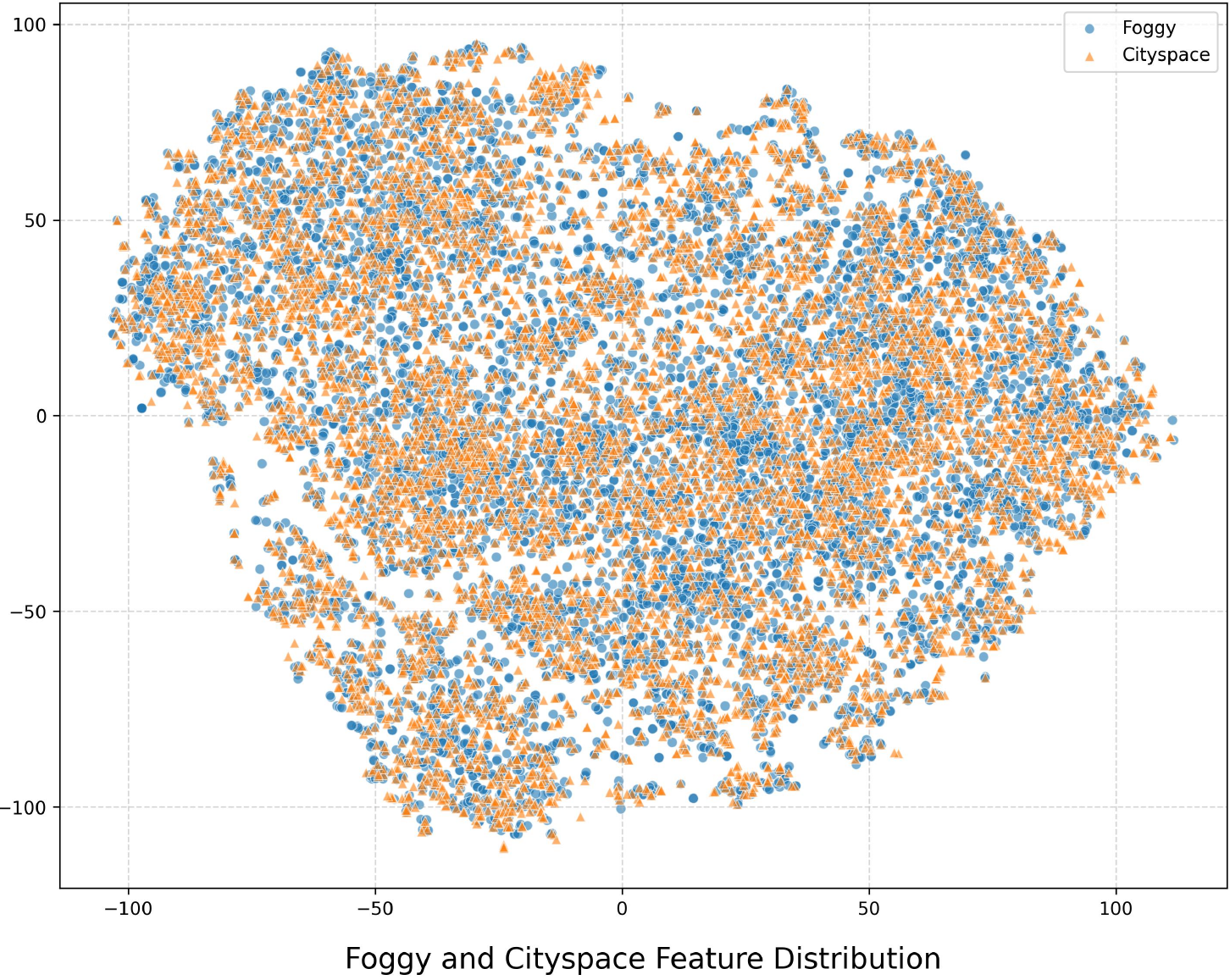}%
\label{fig:a}}
\subfloat[]{\includegraphics[width=2.3in]{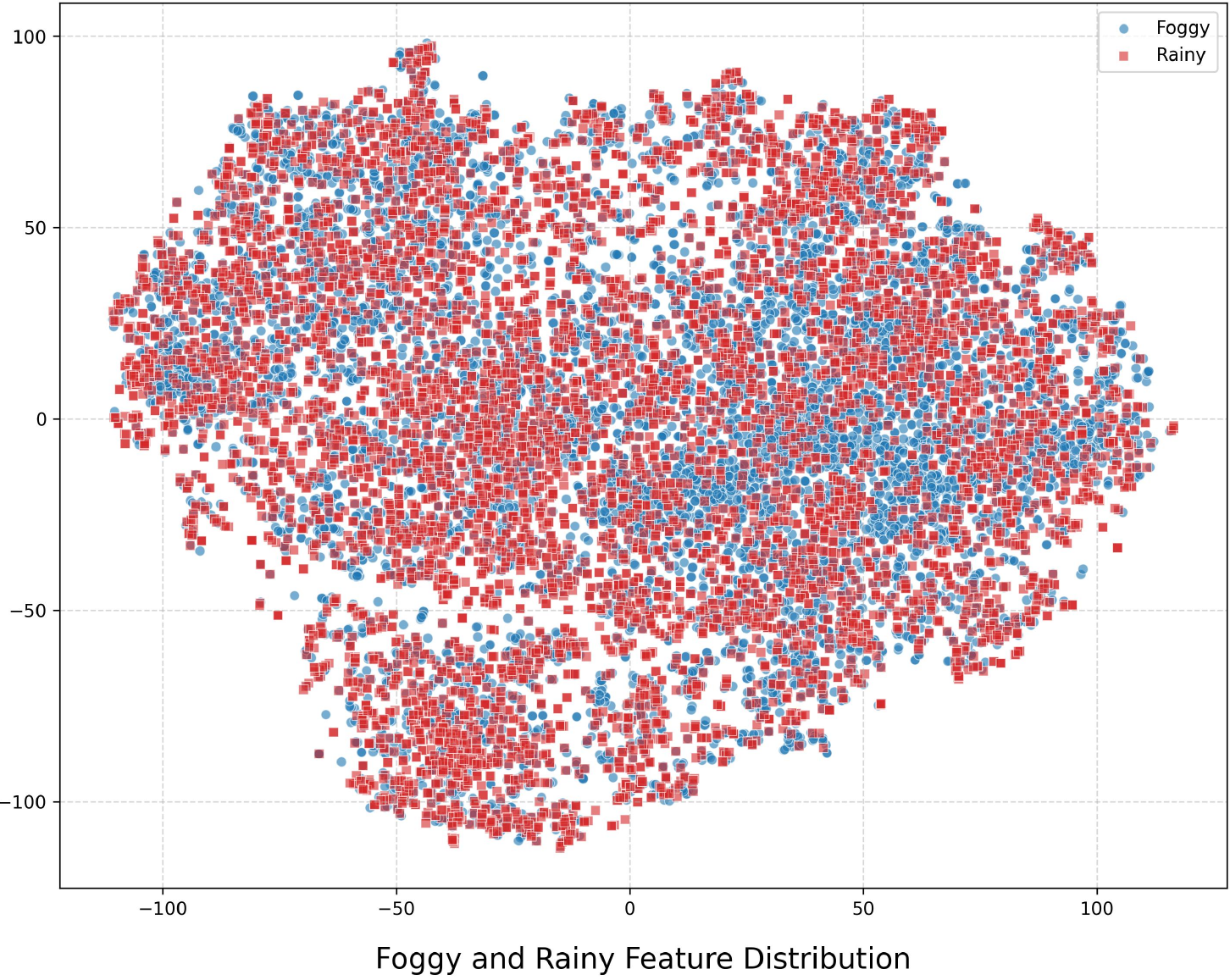}%
\label{fig:b}}
\subfloat[]{\includegraphics[width=2.3in]{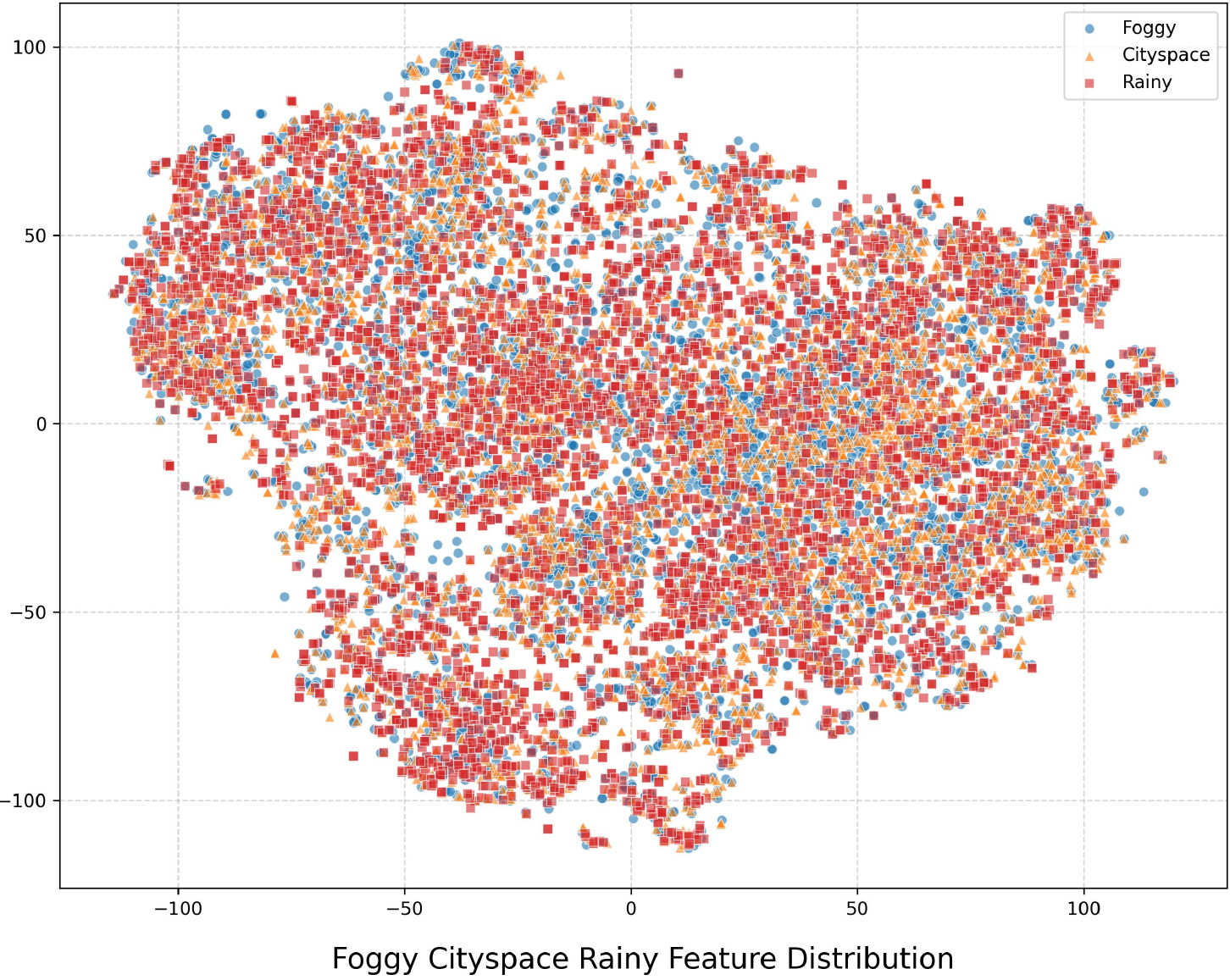}%
\label{fig:c}}
\caption{Analysis of Differences in the Distribution of Features. Where Foggy represents the source domain, Cityspace and Rainy represent different target domains, respectively}
\label{figs:didm}
\end{figure*}

\begin{figure}
    \centering
    \includegraphics[width=0.95\linewidth]{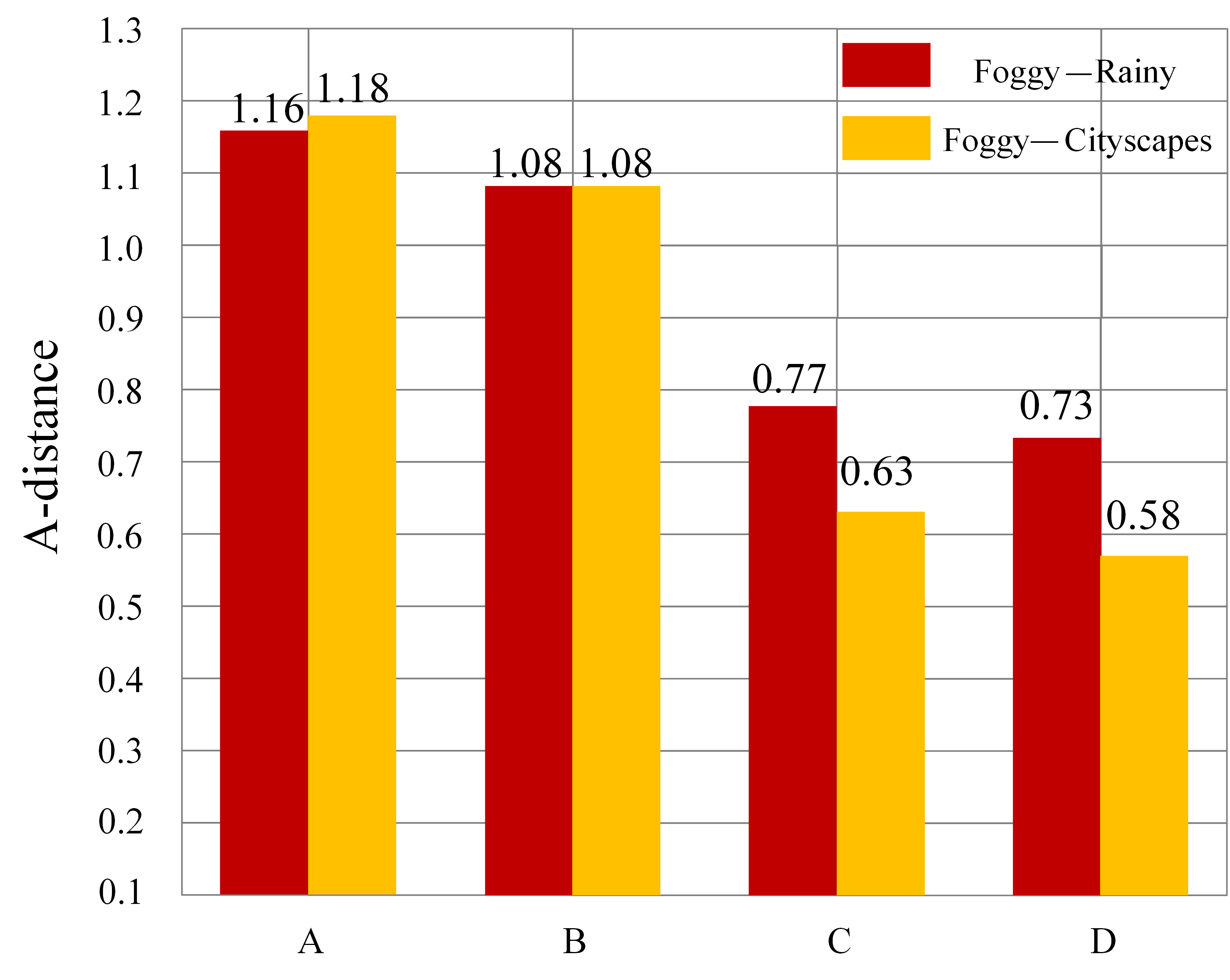}
    \caption{A-distance Result Display. Where A represents the absence of \textbf{FD} and \textbf{$\beta$}; B indicates the absence of \textbf{FD}; C signifies the absence of \textbf{$\beta$}; and D denotes full \textbf{DIDM}.}
    \label{fig:A-distance}
\end{figure}
\subsection{Hyperparameter Analysis}
Appropriate hyperparameter settings are essential for optimizing model performance.

First, in order to effectively differentiate between individual domain-specific features while retaining meaningful domain-specific information, the selection of hyperparameter $\tau$ in Eq. \ref{eq:pythagoras7} is crucial. As shown in Fig. \ref{fig:short-b}, the model demonstrates the best generalization performance across multiple datasets when $\tau$ = 0.1. This clearly demonstrates that the appropriate hyperparameter A can effectively balance feature differentiation and information retention, thereby enhancing the model's adaptability and robustness across various domains.

Second, to verify the effectiveness of the dynamic weighting parameter $\beta$ in Eq. \ref{eq:pythagoras10}, we conducted a comparative experiment. One group utilized parameter $\beta$ when the features were aligned, while the other group did not. The experimental results, illustrated in Fig. \ref{fig:short-c}, indicate that the use of weighting parameter $\beta$ effectively prevents the over-alignment of features, thereby better maintaining feature diversity. This characteristic significantly enhances the model's generalization ability and robustness across different datasets, clearly demonstrating the crucial role of parameter $\beta$ in optimizing the feature alignment process.

Last, as shown in Fig. \ref{fig:short-a}, we experimented with the hyperparameters $\alpha$ in Eq. \ref{eq:pythagoras12}. The results indicate that the selection of hyperparameters significantly impacts model performance, with the model achieving optimal results across different datasets when parameter $\alpha$ = 0.45 is utilized.

\subsection{Analysis of Differences in the Distribution of Features}
The visualization results of the feature distributions for the source and target domains, as illustrated in Fig. \ref{figs:didm} and Fig. \ref{fig:A-distance}, demonstrate significant overlapping regions between the feature distributions of the two domains. This finding confirms that the DIDM can effectively balance domain-invariant and domain-specific features, thereby reducing inter-domain distribution differences while preserving the discriminative features in cross-domain tasks. The A-distance for Cityspace from Foggy and Rainy conditions is 0.58 and 0.73, respectively.

\section{Conclusion}
\label{sec:conclusion}
Existing domain-generalized object detection models (S-DGOD) primarily focus on learning domain-invariant features to mitigate the negative impact of domain bias on the model's generalization ability. However, this approach often overlooks the inherent differences between various domains and may complicate the training process and lead to a loss of valuable information. To address this issue, we propose a Diversity Invariant Detection Model (DIDM) in this paper. This approach aims to balance domain invariance and domain diversity, thereby enhancing the model's feature extraction capabilities. In DIDM, we propose to introduce the Diversity Learning Module (DLM) and Weighted Aligning Module (WAM) for the domain-specific feature and domain invariance, respectively. For DLM, the feature diversity (FD) loss is implemented with entropy maximization loss to eliminate the semantic information while keeping the feature diversity. Additionally, a Weighted Aligning Module (WAM) is incorporated to prevent the overemphasis on the feature alignment with loss weight. With the combination of the DLM and WAM, the DIDM can efficiently manage the domain-specific and domain-invariance when preserving the feature diversity, resulting in improved detection performance across multiple domains. This strategy not only enhances the model's adaptability to domain variations but also ensures stable detection performance in diverse environments. Both experimental data and comprehensive analysis validate the effectiveness of DIDM.



{
    \small
    \bibliographystyle{IEEEtran}
    \bibliography{main}
}

\newpage

\vfill

\end{document}